\documentclass[]{TEAI}
\usepackage{helvet}

\usepackage{amsmath} 
\usepackage{natbib}
\usepackage{graphicx}
\usepackage{subcaption} 

\usepackage[toc,page,header]{appendix}
\usepackage[utf8]{inputenc} 
\usepackage[T1]{fontenc}    
\usepackage{hyperref}       
\usepackage{url}            
\usepackage{booktabs}       
\usepackage{lmodern}        
\usepackage{amsfonts}       
\usepackage{nicefrac}       
\usepackage{microtype}      
\usepackage{wrapfig}

\usepackage{amssymb}  
\usepackage{fontawesome}  
\usepackage{url}  

\usepackage{titletoc}

\usepackage{tikz}  
\usepackage{comment}  
\usepackage{tabularx}  
\usepackage{booktabs}  

\usepackage{minitoc}

\usepackage{booktabs}
\usepackage{array}
\usepackage{etoolbox}

\definecolor{lightblue}{RGB}{200, 230, 255}  
\definecolor{headerblue}{RGB}{150, 200, 255} 

\usepackage{pgfplots}
\usepackage[utf8]{inputenc} 
\usepackage[T1]{fontenc}   
\usepackage{hyperref}       
\usepackage{url}            
\usepackage{booktabs}       
\usepackage{amsfonts}       
\usepackage{nicefrac}       
\usepackage{microtype}      
\usepackage{xcolor}         
\usepackage{graphicx}
\usepackage{float}
\usepackage{comment}
\usepackage{multirow} 
\usepackage{amsmath} 
\usepackage{makecell} 
\usepackage{siunitx}  
\usepackage{tikz}
\usepackage{pgf-pie} 
\usepackage{subcaption}
\usepackage{wrapfig}
\usepackage[export]{adjustbox}

\usepackage{ragged2e}      
\usepackage{tabularx}       
\usepackage{array}          
\usepackage{caption}     
\usepackage{enumitem}
\usepackage{pifont}
\usepackage[hang,flushmargin]{footmisc} 

\usepackage{tcolorbox}

\usepackage{tcolorbox}    
\tcbuselibrary{breakable}  
\tcbuselibrary{skins}      

\usepackage{tabularx}
\usepackage{listings}

\def\eg{e.g.} 

\def\ie{i.e.}

\title{VideoLoom: A Video Large Language Model for Joint Spatial-Temporal Understanding}

\author{
    Jiapeng Shi\textsuperscript{1}, 
    Junke Wang\textsuperscript{1}, 
    Zuyao You\textsuperscript{1}, 
    Bo He\textsuperscript{2}, 
    Zuxuan Wu\textsuperscript{1,$\dagger$}
}

\affiliation[1]{\mbox{Fudan University}} 
\affiliation[2]{\mbox{University of Maryland, College Park}}

\abstract{
\begin{abstract}

This paper presents VideoLoom, a unified Video Large Language Model (Video LLM) for joint spatial-temporal understanding. To facilitate the development of fine-grained spatial and temporal localization capabilities, we curate LoomData-8.7k, a human-centric video dataset with temporally grounded and spatially localized captions. With this, VideoLoom achieves state-of-the-art or highly competitive performance across a variety of spatial and temporal benchmarks (\eg, 63.1 \begin{math} \mathcal{J\&F} \end{math} on ReVOS for referring video object segmentation, and 48.3 R1@0.7 on Charades-STA for temporal grounding). In addition, we introduce LoomBench, a novel benchmark consisting of temporal, spatial, and compositional video–question pairs, enabling a comprehensive evaluation of Video LLMs from diverse aspects. Collectively, these contributions offer a universal and effective suite for joint spatial-temporal video understanding, setting a new standard in multimodal intelligence.
\end{abstract}
}

\correspondence{\email{zxwu@fudan.edu.cn}}
\checkdata[Website]{\url{https://github.com/JPShi12/VideoLoom}}

\begin{document}
\maketitle
\renewcommand{\thefootnote}{}
\footnotetext{$^\dagger$Corresponding authors.}
\renewcommand{\thefootnote}{\arabic{footnote}}

\vspace{-1.5em}

\section{Introduction}
\label{sec:intro}
Recent years have witnessed the rapid development of Multimodal Large Language Models (MLLMs)~\cite{hurst2024gpt,team2024gemini,bai2023qwen,bai2025qwen2, chen2024expanding}, extending their scope from static image understanding~\cite{liu2023visual,liu2024improved,wang2023see,meng2024deepstack,chen2025comp} to dynamic video comprehension~\cite{li2023videochat,maaz2023video,wang2023chatvideo,zhang2023video,li2024llama,peng2024inst}. Video Large Language Models (Video LLMs), which integrate spatial perception with temporal reasoning, have demonstrated strong generalization and competitive performance across a wide range of multimodal benchmarks. More recently, increasing efforts have been devoted to equipping Video LLMs with fine-grained understanding capabilities, such as temporal grounding~\cite{ren2024timechat,huang2024vtimellm,wang2024hawkeye,guo2025vtg,li2025universal}, referring video segmentation~\cite{yan2024visa,gong2025devil,lin2025glus,yuan2025sa2va}, and object tracking~\cite{zhu2023tracking,bai2024one,yang2024samurai}. Despite these achievements, most existing models still focus on either temporal or spatial dimension in isolation, limiting their ability to holistically interpret complex spatial-temporal events in real-world scenarios.

While joint spatial-temporal understanding represents a promising direction for Video LLMs, several critical challenges still remain. First and foremost, a fundamental limitation is the scarcity of high-quality datasets with fine-grained spatial-temporal annotations. Most existing datasets provide either temporal (\eg, event segments~\cite{krishna2017dense,zhou2018towards}) or spatial labels (\eg, object trajectories~\cite{ding2023mevis,seo2020urvos}), but rarely both. A straightforward practice is to jointly train on both types of datasets, but inconsistencies in annotation formats and data distributions often lead to unstable training and hinder the model from establishing coherent spatial–temporal associations. In addition, spatial and temporal video tasks inherently demand different input granularities, \ie, spatial tasks typically require higher resolutions to capture fine-grained details~\cite{yan2024visa,yuan2025sa2va}, while temporal tasks depend on denser frame sampling to model motion dynamics~\cite{ren2024timechat,guo2024trace}. Under fixed computational budgets, it is difficult to balance both requirements, making joint spatial–temporal modeling within a single framework inherently challenging.

To address the above issues, we first introduce LoomData-8.7k, a novel dataset with consistent spatial and temporal annotations. LoomData-8.7k sources videos from ActivityNet~\cite{caba2015activitynet} and is annotated using an automatic pipeline. Specifically, we first segment each untrimmed video into multiple shots and then identify the main characters in the initial shot. Based on this, we track trajectories and generate corresponding action descriptions for each character. This character-centric, shot-guided automatic annotation pipeline provides richer spatial references and complete temporal coverage, enabling detailed and coherent spatial–temporal understanding.

With this, we further introduce VideoLoom, a simple yet effective video large language model (Video LLM) for joint spatial–temporal understanding. To accommodate both capabilities within a single framework, we combine multi-frame inputs that capture temporal dynamics with high-resolution keyframe inputs that preserve fine-grained spatial details. Two types of visual tokens, \ie, fast tokens and slow tokens, are introduced to balance temporal coverage and spatial precision. The former are generated from up to 128 frames uniformly sampled across the entire video span, providing global temporal context with a low token density per frame. The latter are extracted from 5 keyframes, each allocated a higher token density to encode spatial details at high resolution. These SlowFast visual tokens are interleaved with language instructions to form the input sequence of the Video LLM, enabling coherent and efficient spatial–temporal reasoning over the entire video.

To comprehensively evaluate the spatial–temporal understanding capability of Video LLMs, we also propose LoomBench, a benchmark comprising 130 videos and over 1,400 question-answering pairs spanning temporal grounding and spatial segmentation. Unlike existing datasets that assess these dimensions separately~\cite{krishna2017dense, ding2023mevis}, LoomBench consists of carefully designed questions that require models to perform grounding and segmentation simultaneously. 

Experiment results demonstrate that VideoLoom achieves new state-of-the-art on a wide range of video understanding benchmarks, including spatial benchmarks (\eg, 51.7 \begin{math} \mathcal{J\&F} \end{math} on MeVIS~\cite{ding2023mevis}, 63.1 \begin{math} \mathcal{J\&F} \end{math} on ReVOS~\cite{yan2024visa}) and temporal ones (\eg, 48.3 R1@0.7 on Charades-STA~\cite{gao2017tall}, 7.3 SODA\_c on YouCook2~\cite{zhou2018towards}, 63.3 HIT@1 on QVHighlights~\cite{lei2021detecting}). Comparisons with existing Video LLMs on LoomBench further validate the effectiveness of VideoLoom in unified spatial-temporal comprehension.

\vspace{-0.05in}
\section{Related Work}

\vspace{-0.05in}
\subsection{Spatial-Temporal Video Datasets}
Existing video datasets for spatial-temporal understanding can generally be categorized into two separate types: temporal-focused and spatial-focused. The temporal-focused datasets, \eg, for dense captioning~\cite{zhou2018towards,krishna2017dense} or temporal grounding~\cite{gao2017tall}, provide descriptions or queries aligned with timestamps but typically lack spatial annotations. In contrast, the spatial-focused datasets focus on spatial localization through segmentation masks~\cite{khoreva2019video,seo2020urvos,ding2023mevis,yan2024visa} or trajectory annotations~\cite{fan2019lasot, muller2018trackingnet, huang2019got}, but do not include detailed temporal locations of actions. Few datasets focus on atomic actions, featuring coarse-grained spatial-temporal tubelets~\cite{gu2018ava, zhang2020does}, yet constrained by extremely short durations, typically around 10 seconds. Additionally, current datasets rely on costly manual annotations with brief captions of objects or events, lacking detailed positional references and temporal coverage. Collectively, these factors hinder the training of Video LLMs with spatial-temporal comprehension. To bridge this gap, we introduce LoomData-8.7k, providing both fine-grained temporal annotations and mask-level spatial tracklets for long-form videos at scale, enabling more comprehensive spatial-temporal modeling.

\vspace{-0.1in}
\subsection{Video Large Language Models}
Recent advancements in MLLMs reveal a clear trend in visual comprehension from basic image~\cite{liu2023visual, meng2024deepstack, you2025pix2cap} to video stream~\cite{wang2024omnivid,he2024ma,li2024llama,yuan2025videorefer,han2025videoespresso}. While early models primarily focus on coarse-grained tasks such as captioning and retrieval~\cite{shvetsova2024howtocaption,xu2024carebench}, there is a growing need for fine-grained understanding that captures precise object interactions and temporal dynamics. Within this landscape, Video LLMs designed for fine-grained video understanding can be broadly categorized into two directions: temporal-focused and spatial-focused models. The former, such as TimeChat~\cite{ren2024timechat} and TRACE~\cite{guo2024trace}, are trained on timestamp-aware instruction data to develop the temporal localization capabilities. Spatial models, on the other hand, focus on grounding visual regions in the format of trajectories~\cite{yan2024visa,yuan2025sa2va,gong2025devil}. While both directions address critical aspects of video understanding, neither is sufficient in isolation. Some works~\cite{xu2024slowfast,huang2024lita,li2025llava,wang2025spacevllm} begin to model the spatial-temporal clues in video simultaneously, yet they remain confined to specific tasks or coarse-grained perception (\eg, sparse spatial bounding boxes). In this paper, we propose a unified Video LLM, VideoLoom, that accommodates both fine-grained temporal understanding and spatial perception within a single framework. 

\vspace{-0.05in}
\section{Method}
\label{sec:Method}
\vspace{-0.05in}
This section introduces the VideoLoom suite, which advances joint spatial-temporal understanding from three key perspectives: 1) \textbf{LoomData-8.7k}, a video dataset with fine-grained spatial and temporal annotations. 2) \textbf{VideoLoom}, a Video LLM that handles joint temporal understanding and spatial perception tasks within a single framework. and 3) \textbf{LoomBench}, a video benchmark developed to evaluate the joint spatial-temporal capability of Video LLMs.

\vspace{-0.05in}
\subsection{LoomData-8.7k}
\label{dataset}
\begin{figure*}[t!]
\centering
\includegraphics[width=\textwidth]{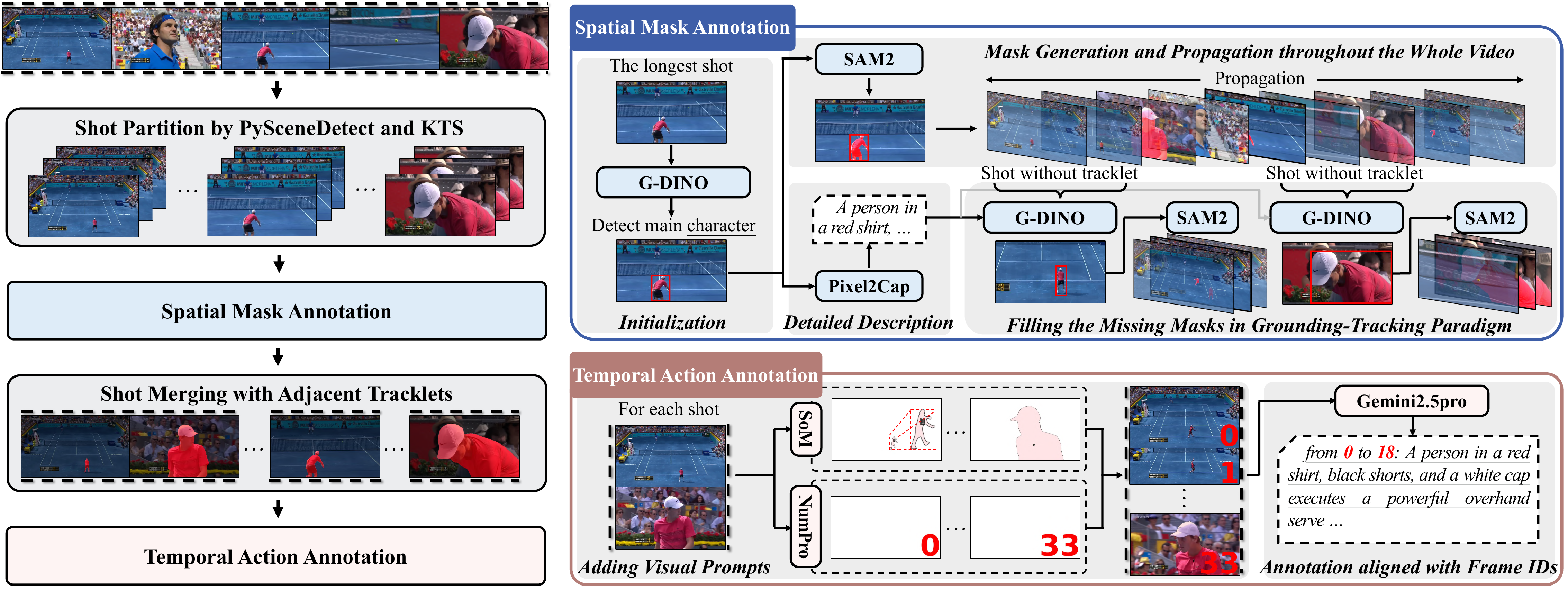}
\vspace{-0.2in}
\caption{Illustration of the designed data annotation pipeline, comprising four stages: shot partition, spatial mask annotation, shot merging, and temporal action annotation. During spatial mask annotation, main characters and their complete tracklets are identified. In temporal action annotation, actions of characters are temporally grounded with visual prompts.}
\label{fig:pipeline}
\vspace{-0.1in}
\end{figure*}

We develop an automatic annotation pipeline that leverages multiple visual foundation models to detect and associate the temporal actions and spatial locations of main characters. As shown in Fig.~\ref{fig:pipeline}, the pipeline comprises four main stages: (i) shot partition, (ii) spatial mask annotation, (iii) shot merging, and (iv) temporal action annotation. 

\noindent\textbf{Shot Partition.} We first partition each video into several shots using PySceneDetect~\cite{Castellano_PySceneDetect} and KTS~\cite{potapov2014category}. PySceneDetect identifies scene boundaries by detecting scene changes between adjacent frames, while KTS captures event transitions. We combine both by sequentially ordering the timestamps of all transition points to achieve accurate shot partition for different scenes. A simple filtering strategy is then applied by merging shots shorter than \(1\) second and discarding videos exceeding \(10\) shots.

\noindent\textbf{Spatial Mask Annotation.} For each video, we use GroundingDINO~\cite{liu2024grounding} to detect the ``person'' category in the center frame of the longest shot, and only keep the bounding box with the highest score as the main character. With this region box, a detailed description of its appearance (\eg, clothing and attributes) is generated by Pix2Cap~\cite{you2025pix2cap}. After this, we employ SAM2~\cite{ravi2024sam} to track the main character throughout the video to produce the initial mask tracklet. We then complete the cross-shot tracklet in a grounding-tracking paradigm. GroundingDINO is also applied to re-annotate this character in the center frame of shots without a tracklet, based on the description. SAM2 then conducts mask tracking on these shots to fill the missing masks, yielding a complete tracklet of the main character. Finally, we perform a manual verification step to refine redundantly tracked shots and remove incorrectly tracked videos. 

\noindent\textbf{Shot Merging.} As a temporal event may span multiple shots (\eg, different camera angles), we merge all adjacent shots annotated with tracklets to obtain temporally consistent annotations. 

\noindent\textbf{Temporal Action Annotation.} With the merged shots and dense trajectories, we then generate detailed, timestamp-aligned action descriptions for main characters in each video. Specifically, we place unique numerical IDs on video frames sampled at 2 FPS in the manner of NumPro~\cite{wu2025number}, and then employ Set-of-Marks (SoM)~\cite{yang2023set} to overlay an instance ID directly onto the segmentation masks of main characters. These sampled frames, along with both visual prompts, are fed into Gemini2.5pro~\cite{comanici2025gemini} to produce fine-grained action descriptions aligned with the frame IDs. 

We annotate the training set of ActivityNet~\cite{caba2015activitynet} with the above pipeline, resulting in 8,710 shots featuring both timestamp-aligned action descriptions and dense spatial masks. On average, each video has a duration of 102.2 seconds and includes 6.0 shots, while the temporal descriptions average 41.3 words. For additional statistics, please refer to Sec.~\ref{section:statistic_loomdata}.

\vspace{-0.05in}
\subsection{VideoLoom}

\begin{figure*}[t!]
\centering
\includegraphics[width=\textwidth]{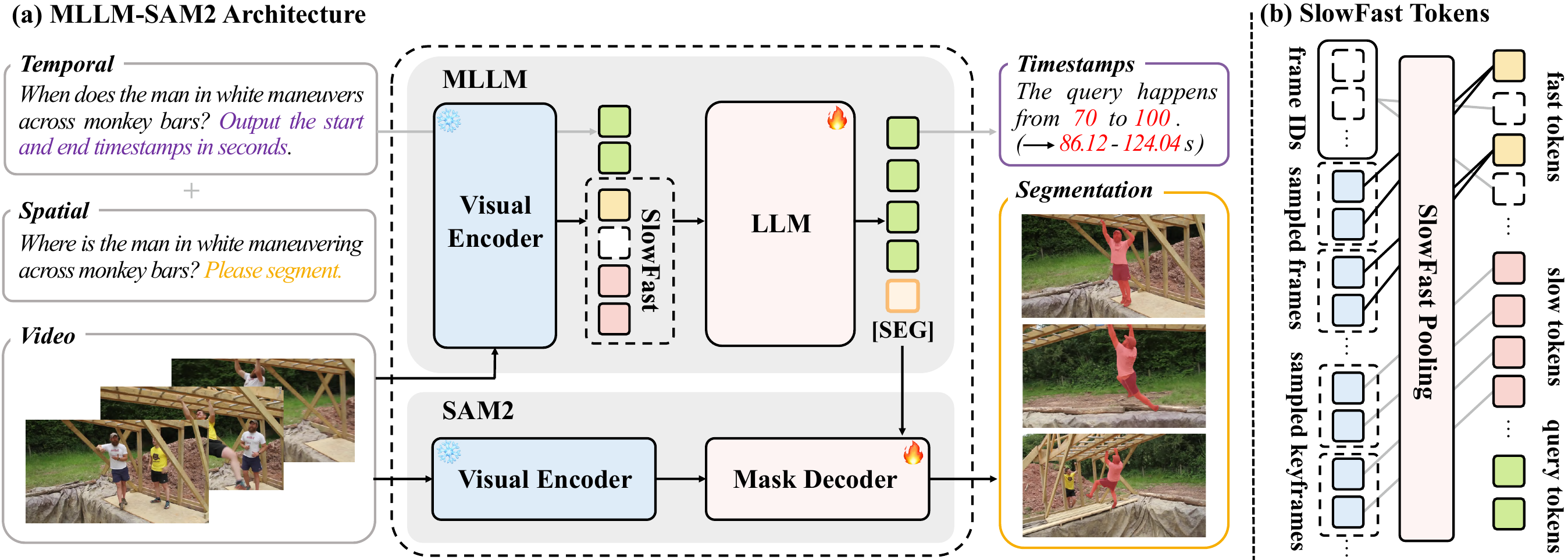}
\vspace{-0.2in}
\caption{Overview of VideoLoom Architecture. Two key designs are: (a) MLLM-SAM2 Architecture, where MLLM and SAM2 are connected via a \texttt{[SEG]} token, unifying temporal understanding and spatial perception. (b) SlowFast Tokens, where input videos are encoded as SlowFast visual tokens to model spatial-temporal representations.}
\label{fig:model}
\vspace{-0.1in}
\end{figure*}

With the above dataset, we further propose VideoLoom, a unified Video LLM to unlock joint spatial-temporal understanding capabilities. Specifically, taking a language query $T$ and a video consisting of $N$ frames $V \in \mathbb{R}^{N \times H \times W \times 3}$ as input, where $H$ and $W$ denote the height and width of each frame respectively, VideoLoom aims to generate an answer text $O$ that contains the required timestamp information, or predict a trajectory in the format of segmentation masks $M \in \mathbb{R}^{N \times H \times W}$:
\begin{equation}
O, M = \mathrm{VideoLoom}(T, V).
\end{equation}

Below, we introduce the SlowFast visual tokens which capture spatial–temporal information at different granularities in Sec.~\ref{slowfast}, the MLLM-SAM2 architecture which integrates these tokens for unified spatial–temporal modeling in Sec.~\ref{mllm_sam2}, and the loss functions in Sec.~\ref{loss}.

\subsubsection{SlowFast Visual Tokens}
\label{slowfast}
Temporal understanding typically requires processing a large number of frames~\cite{ren2024timechat,huang2024vtimellm}, whereas spatial perception demands higher-resolution inputs~\cite{yuan2025sa2va}. To accommodate both, we introduce two types of visual tokens, \ie, fast tokens and slow tokens, which respectively encode dense low-resolution frames with temporal bindings and sparse high-resolution keyframes with rich spatial details.

Specifically, we sparsely sample $N_{s}$ high-resolution keyframes and assign $C$ tokens for each frame to form $N_{s} \times C$ slow tokens. Meanwhile, we also densely sample $N_{f}$ frames across the entire video. Both are fed to a visual encoder~\cite{chen2024expanding} to obtain $N_{s} \times C$ slow tokens and $N_{f} \times \frac{C}{R^2}$ fast tokens, where $R$ denotes the spatial downsampling ratio.

\subsubsection{MLLM-SAM2 Architecture}
\label{mllm_sam2}
\noindent\textbf{Overview.} We integrate InternVL3~\cite{zhu2025internvl3}, a multimodal large language model (MLLM), with SAM2~\cite{ravi2024sam}, a video segmentation and tracking model, to support both spatial and temporal tasks within a unified framework. InternVL3 takes SlowFast visual tokens and text prompts as inputs, producing text responses, timestamps, and a \texttt{[SEG]} token embedding. SAM2 then utilizes this \texttt{[SEG]} token to generate corresponding segmentation masklets. The overall architecture is illustrated in Fig.~\ref{fig:model}.

\noindent\textbf{MLLM for temporal understanding tasks.} 
Our MLLM consists of a visual encoder, a visual projection layer, and an LLM. The sampled frames are input to the visual encoder and then mapped into visual tokens by the visual projection layer. Unlike previous work using absolute timestamps~\cite{ren2024timechat, zeng2024timesuite} or special time tokens~\cite{huang2024lita, guo2024trace}, we interleave unique frame IDs between visual tokens to indicate temporal order. The complete token sequence is used as input to the LLM, which models the spatial-temporal visual features and generates text token predictions according to text queries. Note that for timestamp-related queries, the LLM outputs corresponding frame IDs in text responses to indicate temporal locations.

\noindent\textbf{SAM2 for spatial understanding tasks.}
Given the keyframes sampled for slow tokens, we input them to SAM2 to predict spatial trajectories. A \texttt{[SEG]} token is used to connect MLLM with SAM2 mask decoder, providing the mask decoder with rich target information and prompting it to generate masks in the keyframes. We then propagate these masks to the entire video using a visual memory~\cite{yuan2025sa2va}.

\subsubsection{Loss Functions}
\label{loss}

VideoLoom is trained in an end-to-end manner with the following objective:
\begin{equation}
\mathcal{L} = \lambda_{\text{text}}\mathcal{L}_{\text{text}} + \lambda_{\text{mask}}\mathcal{L}_{\text{mask}},
\end{equation}

where $\mathcal{L}_{\text{text}}$ denotes the standard cross-entropy loss for text generation, and $\mathcal{L}_{\text{mask}}$ indicates the segmentation loss combining per-pixel binary cross-entropy (BCE) loss and DICE loss~\cite{milletari2016v}. $\lambda_{\text{text}}$ and $\lambda_{\text{mask}}$ are balancing hyper-parameters.

\subsection{LoomBench}
\label{sec:loombench}

\begin{figure*}[t!]
\centering
\includegraphics[width=\textwidth]{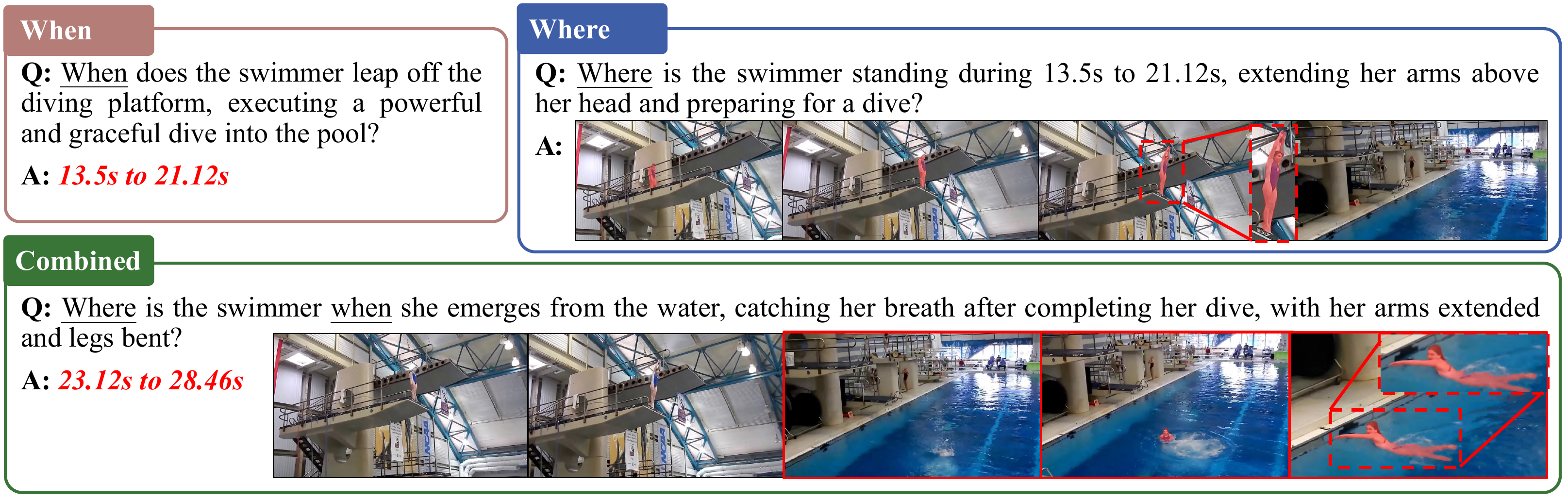}
\vspace{-0.2in}
\caption{Visualization of the QA pairs in LoomBench. Three types of QA are shown: \textit{When} targets the action timestamps given a query and the whole video, \textit{Where} targets the person masklet given a query and a certain video segment, while \textit{Combined} directly targets the tracklet segment corresponding to the query.}
\label{fig:bench}
\vspace{-0.1in}
\end{figure*}

We curate LoomBench, a new benchmark designed to jointly evaluate the spatial and temporal understanding capabilities of Video LLMs. Specifically, we apply the automatic annotation pipeline described in Sec.~\ref{dataset} to the validation set of ActivityNet~\cite{caba2015activitynet} to generate preliminary annotations. These annotations are then manually verified and refined to further improve quality and consistency. For each video shot, we prompt LLaMA3.1~\cite{grattafiori2024llama} to generate three types of questions based on the action descriptions of the main characters: \textit{When}, \textit{Where}, and \textit{Combined}. As a result, LoomBench contains 130 videos, with an average of 4.2 temporal shots per video and an average shot length of 17.6 seconds. A visualization example is shown in Fig.~\ref{fig:bench}.

\noindent \textbf{\textit{When}/\textit{Where}} questions respectively target the action timestamps and the person masks of each segment, focusing on the evaluation of temporal understanding and spatial perception. Following existing benchmarks~\cite{gao2017tall,krishna2017dense}, we adopt R1@0.5 and temporal IoU (tIoU) as evaluation metrics for \textit{When} questions. For \textit{Where} questions, we use the $\mathcal{J\&F}$ metric~\cite{seo2020urvos,ding2023mevis}, which averages region similarity ($\mathcal{J}$) and contour accuracy ($\mathcal{F}$). LoomBench contains 541 \textit{When} and 487 \textit{Where} questions.

\noindent\textbf{\textit{Combined}} questions such as ``Where is the person when he/she is doing something?'' extend beyond the scope of existing datasets and enable more comprehensive evaluation of unified spatial–temporal understanding. We annotate 456 \textit{Combined} questions in LoomBench. The standard $\mathcal{J\&F}$ metric computes the difference between predicted masklets and groundtruth across all video frames. However, for \textit{Combined} questions, the duration of the queried tracklet constitutes only a small fraction of the entire video (on average, 20.9\%), making $\mathcal{J\&F}$ dominated by background frames without mask annotations. These backgrounds can inflate $\mathcal{J\&F}$ scores and undermine their reliability for evaluation. To address this issue, we propose Bidirectional Foreground $\mathcal{J\&F}$, which computes $\mathcal{J\&F}$ within the temporal intervals of both the predicted and groundtruth foreground masks, and then takes their harmonic mean:
\begin{align}
\mathcal{J\&F}_{bi\text{-}fore} &=
\frac{(\mathcal{J}_p+\mathcal{F}_p)\times (\mathcal{J}_g+\mathcal{F}_g)}
{(\mathcal{J}_p+\mathcal{F}_p)+(\mathcal{J}_g+\mathcal{F}_g)}
\label{eq:bi-fore}
\end{align}
where $\mathcal{J}_p = \mathcal{J}_{\mathrm{Loc}(P)}(P, G)$, $\mathcal{J}_g = \mathcal{J}_{\mathrm{Loc}(G)}(P, G)$, $\mathcal{F}_p = \mathcal{F}_{\mathrm{Loc}(P)}(P, G)$, and $\mathcal{F}_g = \mathcal{F}_{\mathrm{Loc}(G)}(P, G)$.  \begin{math} P \end{math}, \begin{math} G \end{math} denote the predicted and groundtruth masks, and the function \begin{math} \mathrm{Loc} \end{math} extracts the temporal span of a masklet. Accordingly, \begin{math} \mathcal{J}_p \end{math} refers to the \begin{math} \mathcal{J} \end{math} score computed over the temporal segment of predicted masklet, and so on. For more analysis on $\mathcal{J\&F}_{bi\text{-}fore}$, please refer to Sec.~\ref{section:bf_jf}.

\vspace{-0.05in}
\section{Experiments}
\subsection{Experimental Setup}
\label{sec:Implementation}
\noindent \textbf{Training data:} Our training data can be categorized into four types: 1) image question answering (QA), which includes LLaVA-665k~\cite{liu2024improved}. 2) image segmentation data, comprising standard referring expression segmentation datasets~\cite{kazemzadeh2014referitgame, yu2016modeling} and grounding conversation generation (GCG) data~\cite{rasheed2024glamm}. 3) video segmentation data, including RefYTVOS~\cite{seo2020urvos}, MeVIS~\cite{ding2023mevis}, and ReVOS~\cite{yan2024visa}. 4) video temporal instruction data, consisting of Charades-STA~\cite{gao2017tall}, YouCook2~\cite{zhou2018towards}, and QVHighlights~\cite{lei2021detecting}. The proposed LoomData-8.7k is converted into both referring video object segmentation (VOS) and temporal grounding formats for joint training.

\noindent \textbf{Implementation details:} We choose InternVL3~\cite{zhu2025internvl3} as our foundation MLLM and SAM2~\cite{ravi2024sam} as the segmentation module. A special token \texttt{[SEG]} is added for the mask generation following LISA~\cite{lai2024lisa}. The input frames are resized to 448$\times$448 and 1024$\times$1024 for the MLLM and SAM2 visual encoders, respectively. The number of slow visual tokens $C$ per frame is set to 256, and the downsampling ratio $R$ is kept at 4, resulting in 16 fast tokens per frame. Up to 128 frames are uniformly sampled for fast tokens, while only 5 keyframes are encoded as slow tokens. We use the XTuner~\cite{contributors2023xtuner} codebase for training and evaluation, finetuning only the mask decoder and LLM module while keeping the visual encoder frozen. The LLM is adapted via LoRA~\cite{hu2022lora}, with a learning rate of \(4\times10^{-5}\). The loss weights $\lambda_{\text{text}}$ and $\lambda_{\text{mask}}$ are both set to 1. We train VideoLoom for one epoch with a global batch size of 64. All experiments are facilitated on 8 NVIDIA H20 GPUs with 96 GB of memory.

\subsection{Main Results}

\definecolor{gray}{HTML}{808080}
\begin{table*}[t!]
\makebox[\linewidth][s]{%
\begin{minipage}[t]{0.56\linewidth}
\centering
\small
\setlength{\tabcolsep}{3pt} 
\renewcommand{\arraystretch}{1.28425}
\begin{tabular}{l|cc|ccc|cc}
\toprule
\multicolumn{1}{l|}{\multirow{2}{*}{\textbf{Method}}} & \multicolumn{2}{c|}{\textbf{Charades}} & \multicolumn{3}{c|}{\textbf{YouCook2}} & \multicolumn{2}{c}{\textbf{QVHL}} \\ 
\cmidrule(r){2-3} \cmidrule(r){4-6} \cmidrule(r){7-8}
& R1@0.5 & R1@0.7 & S & C & F1 & mAP & HIT@1 \\
\midrule
TimeChat-7B~\cite{ren2024timechat} & 46.7 & 23.7 & 3.4 & 11.0 & 19.5 & 21.7 & 37.9 \\
VTG-LLM-7B~\cite{guo2025vtg} & 57.2 & 33.4 & 3.6 & 13.4 & 20.6 & 24.1 & 41.3 \\
TRACE-7B~\cite{guo2024trace} & 61.7 & 41.4 & 6.7 & 35.5 & 31.8 & \textbf{31.8} & 51.5 \\
TimeSuite-7B~\cite{zeng2024timesuite} & 67.1 & 43.0 & - & - & - & 27.0 & 55.3 \\
\midrule
\textcolor{gray}{HawkEye-7B\textsuperscript{*}~\cite{wang2024hawkeye}} & \textcolor{gray}{58.3} & \textcolor{gray}{28.8} & \textcolor{gray}{-} & \textcolor{gray}{-} & \textcolor{gray}{-} & \textcolor{gray}{-} & \textcolor{gray}{-} \\
\textcolor{gray}{UniTime-7B\textsuperscript{*}~\cite{li2025universal}} & \textcolor{gray}{75.3} & \textcolor{gray}{56.9} & \textcolor{gray}{-} & \textcolor{gray}{-} & \textcolor{gray}{-} & \textcolor{gray}{-} & \textcolor{gray}{-} \\
\midrule
VideoLoom-8B & \textbf{70.0} & \textbf{48.3} & \textbf{7.3} & \textbf{41.5} & \textbf{33.6} & 27.5 & \textbf{63.3} \\
\bottomrule
\end{tabular}
\vspace{-0.06in}
\caption{Performance comparison on diverse temporal understanding benchmarks, \textsuperscript{*} denotes models specifically designed for TVG.}
\label{tab:temporal_results}
\end{minipage}
\hfill
\begin{minipage}[t]{0.41\linewidth}
\centering
\small
\renewcommand{\arraystretch}{1.2}
\setlength{\tabcolsep}{3.5pt}
\begin{tabular}{l| c c c}
\toprule
\multirow{2}{*}{\textbf{Method}}
& \textbf{MeVIS} & \textbf{YTVOS} & \textbf{ReVOS} \\ 
\cmidrule(r){2-2} \cmidrule(r){3-3} \cmidrule(r){4-4}
& \begin{math} \mathcal{J\&F} \end{math} & \begin{math} \mathcal{J\&F} \end{math} & \begin{math} \mathcal{J\&F} \end{math} \\
\midrule
TrackGPT-7B~\cite{zhu2023tracking} & 40.1 & 56.4 & 43.6 \\
VISA-7B~\cite{yan2024visa} & 43.5 & 61.5 & 46.9 \\
ViLLa-6B~\cite{zheng2025villa} & 49.4 & 67.5 & 57.0 \\
GLUS-7B~\cite{lin2025glus} & 51.3 & 67.3 & 54.9 \\
Sa2VA-8B~\cite{yuan2025sa2va} & 46.9 & 70.7 & 57.6 \\
VRS-HQ-7B~\cite{gong2025devil} & 50.6 & 70.4 & 59.1 \\
VRS-HQ-13B~\cite{gong2025devil} & 50.9 & 71.0 & 60.0 \\
\midrule
VideoLoom-8B & \textbf{51.7} & \textbf{71.3} & \textbf{63.1} \\
\bottomrule
\end{tabular}
\caption{Performance comparison on ref-VOS.}
\label{tab:spatial_results}
\end{minipage}%
}
\vspace{-0.15in}
\end{table*}

\noindent\textbf{Comparison on Temporal Benchmarks.}
We evaluate our model on a wide range of temporal tasks, including temporal video grounding (TVG), dense video captioning (DVC), and video highlight detection (VHD), for a comprehensive assessment of its temporal understanding capabilities.

The comparison with existing Video LLMs is reported in Tab.~\ref{tab:temporal_results}. VideoLoom achieves state-of-the-art or competitive performance across TVG, DVC, and VHD, \eg, 48.3 R1@0.7 on Charades-STA and 63.3 HIT@1 on QVHighlights, surpassing both unified models, \eg, TimeSuite~\cite{zeng2024timesuite}, and task-specific models, \eg, HawkEye~\cite{wang2024hawkeye}. This highlights the strong temporal understanding capabilities of our method. Although VideoLoom lags behind UniTime~\cite{li2025universal} on Charades-STA, we attribute this to the much larger amount of grounding data used in their training and the complex inference procedure involving recursive localization.

\noindent\textbf{Comparison on Spatial Benchmarks.}
For spatial understanding in videos, we evaluate our method on referring Video Object Segmentation (VOS) task on RefYTVOS~\cite{seo2020urvos}, MeVIS~\cite{ding2023mevis}, and ReVOS~\cite{yan2024visa}. \begin{math} \mathcal{J\&F} \end{math} is chosen as the metric. The results in Tab.~\ref{tab:spatial_results} show that VideoLoom even outperforms tracking-oriented Video LLMs on all these benchmarks, achieving 51.7 on MeVIS, 71.3 on RefYTVOS, and 63.1 on ReVOS in terms of \begin{math} \mathcal{J\&F} \end{math}. This superior performance showcases the effectiveness of our method for fine-grained spatial understanding.

\begin{table}[h!]  
    \begin{minipage}{0.4\columnwidth}
        Additionally, we also evaluate VideoLoom on image benchmarks, including RefCOCO~\cite{kazemzadeh2014referitgame}, RefCOCO+~\cite{kazemzadeh2014referitgame}, and RefCOCOg~\cite{yu2016modeling} for referring segmentation, and Grand-f~\cite{rasheed2024glamm} for Grounded Conversation Generation (GCG). We adopt cIoU, AP50, and mIoU as the measurement metrics. The comparison results in Tab.~\ref{tab:is_results} demonstrate that VideoLoom achieves the best results on all datasets, further demonstrating its strong spatial capabilities.
    \end{minipage}
    \hfill  
    \begin{minipage}{0.58\columnwidth}
        \centering
        \small
        \renewcommand{\arraystretch}{1.1}
        \begin{tabularx}{\columnwidth}{l|>{\centering\arraybackslash}X>{\centering\arraybackslash}X>{\centering\arraybackslash}X|>{\centering\arraybackslash}X>{\centering\arraybackslash}X}
        \toprule
        \multicolumn{1}{l|}{\multirow{2}{*}{\textbf{Method}}} & \textbf{RC}   & \textbf{RC+}  & \textbf{RCg}  & \multicolumn{2}{c}{\textbf{GCG}} \\
        \cmidrule(r){2-4} \cmidrule(r){5-6}
        & cIoU & cIoU & cIoU & AP50       & mIoU       \\
        \midrule
        VRS-HQ-7B~\cite{gong2025devil} & 73.5 & 61.7 & 66.7 & - & -    \\
        LISA-7B~\cite{lai2024lisa} & 74.9 & 65.1 & 67.9 & - & -    \\
        OMG-LLaVA-7B~\cite{zhang2024omg} & 78.0 & 69.1 & 72.9 & 29.9 & 65.5 \\
        GLaMM-7B~\cite{rasheed2024glamm} & 79.5 & 72.6 & 74.2 & 30.8 & 66.3 \\
        Sa2VA-8B~\cite{yuan2025sa2va} & 81.6 & 76.2 & 78.7 & 31.0 & -    \\
        \midrule
        VideoLoom-8B & \textbf{83.4} & \textbf{79.2} & \textbf{81.4} & \textbf{34.1} & \textbf{68.6} \\
        \bottomrule
        \end{tabularx}
        \vspace{-0.1in}
        \caption{Performance comparison on image segmentation benchmarks.}
        \label{tab:is_results}
    \end{minipage}
\end{table}

\noindent\textbf{Comparison on LoomBench.}
Finally, we evaluate the joint spatial-temporal comprehension capability on the proposed LoomBench. For comparison, we design a strong baseline that first adopts TimeSuite-7B~\cite{zeng2024timesuite} to localize the relevant clip in the given video, and then applies Sa2VA-8B~\cite{yuan2025sa2va} to segment the masks based on the user query (denoted as TimeSuite + Sa2VA). We incorporate tIoU and \begin{math} \mathcal{J\&F}_{bi\text{-}fore} \end{math} to evaluate \textit{Combined} questions.

\makebox[\linewidth][s]{As shown in Tab.~\ref{tab:loombench_results}, VideoLoom outperforms the above baseline by a clear margin on \textit{Combined} questions}
\vspace{-9pt}
\begin{table}[h!]  
    \begin{minipage}{0.45\columnwidth}
          (+16.2 and +15.4 in terms of tIoU and \begin{math} \mathcal{J\&F}_{bi\text{-}fore} \end{math}). This not only validates the effectiveness of our model on this task, but also underscores the necessity of joint spatial–temporal understanding for comprehensive video comprehension. In addition, we also evaluate VideoLoom on \textit{When} and \textit{Where} questions, demonstrating robust performance in both temporal comprehension and spatial perception.
    \end{minipage}
    \hfill  
    \begin{minipage}{0.53\columnwidth}
        \centering
        \small
        \setlength{\tabcolsep}{4.8pt}
        \renewcommand{\arraystretch}{1.1}
        \begin{tabularx}{\columnwidth}{l | cc | c | c>{\centering\arraybackslash}X }
        \toprule
        \multirow{2}{*}{\textbf{Method}} & \multicolumn{2}{c|}{\textbf{\textit{When}}} & \multicolumn{1}{c|}{\textbf{\textit{Where}}} & \multicolumn{2}{c}{\textbf{\textit{Combined}}} \\
        \cmidrule(r){2-3} \cmidrule(r){4-4} \cmidrule(r){5-6}
        & R1     & tIoU     & \begin{math} \mathcal{J\&F} \end{math}            
        & tIoU               & \begin{math} \mathcal{J\&F}_{bi\text{-}fore} \end{math} \\
        \midrule
        TimeSuite-7B~\cite{zeng2024timesuite} & 23.1 & 27.6 & - & - & - \\
        Sa2VA-8B~\cite{yuan2025sa2va} & -  & -  & 86.1  & - & - \\
        TimeSuite+Sa2VA & - & - & - & 25.4 & 33.7 \\
        VideoLoom-8B & \textbf{37.9} & \textbf{39.7}  & \textbf{87.2}  & \textbf{41.6} & \textbf{49.1}  \\
        \bottomrule
        \end{tabularx}
        \vspace{-0.1in}
        \caption{Performance comparison on LoomBench.}
        \label{tab:loombench_results}
    \end{minipage}
\vspace{-8pt}
\end{table}

\subsection{Ablation Studies}
\label{ablation}

In this section, we conduct extensive ablation experiments using InternVL2.5-4B~\cite{chen2024expanding}, a lightweight MLLM, as our backbone to study the contribution of different components. 

\noindent \textbf{Effects of SlowFast Visual Tokens.} We build different variants to study the effects of SlowFast visual tokens: 1) using only slow tokens to train on spatial tasks, 2) using only fast tokens to train on temporal tasks, 3) using slow or fast tokens and train on both tasks jointly, 4) using fast tokens for temporal and slow tokens for spatial tasks, and 5) using both slow and fast tokens and train on both tasks. Results of all configurations are compared in Tab.~\ref{tab:slowfast_results}. 

\definecolor{lightgray}{HTML}{D9D9D9}
\begin{table*}[h]
\centering
\small
\setlength{\tabcolsep}{3pt} 
\renewcommand{\arraystretch}{1.1}
\begin{tabularx}{\linewidth}{l|ccc|cc|cc|>{\centering\arraybackslash}X|c|ccc}
\toprule
\multirow{2}{*}{\textbf{Setting}} & \multicolumn{3}{c|}{\textbf{Charades}} & \multicolumn{2}{c|}{\textbf{YouCook2}} & \multicolumn{2}{c|}{\textbf{QVHL}} & \textbf{MeVIS} & \textbf{YTVOS} & \multicolumn{3}{c}{\textbf{ReVOS}} \\
\cmidrule(r){2-4} \cmidrule(r){5-6} \cmidrule(r){7-8} \cmidrule(r){9-9} \cmidrule(r){10-10} \cmidrule(r){11-13}
& R1@0.5 & R1@0.7 & mIoU & S & F1 & mAP & mIoU & \begin{math} \mathcal{J\&F} \end{math} & \begin{math} \mathcal{J\&F} \end{math}  & \begin{math} \mathcal{J\&F}_{\mathrm{Ref.}} \end{math} & \begin{math} \mathcal{J\&F}_{\mathrm{Rea.}} \end{math} & \begin{math} \mathcal{J\&F} \end{math} \\ 
\midrule
Spatial (Slow) & - & - & - & - & - & - & - & 46.8 & 69.1 & 62.3 & 56.7 & 59.5 \\
Temporal (Fast) & 66.1 & 41.4 & 55.8 & 6.6 & 30.3 & \textbf{26.8} & 52.4 & - & - & - & - & - \\
Joint (Slow) & 38.8 & 17.7 & 38.6 & 0.8 & 4.8 & 19.1 & 42.2 & 47.4 & 68.7 & 61.8 & 56.0 & 58.9 \\
Joint (Fast) & 63.3 & 39.0 & 54.3 & 6.5 & 28.6 & 26.2 & 54.8 & 44.6 & 66.2 & 60.0 & 53.6 & 56.8 \\
Joint (Slow/Fast) & 62.2 & 39.0 & 54.0 & 6.0 & 26.4 & 24.2 & 47.1 & 47.6 & 68.9 & 61.6 & 56.0 & 58.8 \\
\rowcolor{lightgray}
Joint (SlowFast)  & \textbf{66.2} & \textbf{43.0} & \textbf{56.5} & \textbf{7.0} & \textbf{30.3} & 25.8 & \textbf{57.2} & \textbf{50.0} & \textbf{70.0} & \textbf{62.5} & \textbf{57.6} & \textbf{60.0} \\
\bottomrule
\end{tabularx}
\vspace{-0.06in}
\caption{Ablation experiments on SlowFast visual tokens.}
\label{tab:slowfast_results}
\end{table*}

Using either slow or fast tokens alone leads to substantial performance degradation on spatial or temporal tasks, respectively. The joint (Slow/Fast) setting, which assigns fast tokens for temporal and slow tokens for spatial, yields more balanced results across all datasets, though still with a noticeable drop compared to the specialized single-task models. When SlowFast tokens are employed, the model achieves consistent improvements across nearly all benchmarks, surpassing standalone spatial or temporal models by 4.8 mIoU on QVHighlights and 3.2 \begin{math} \mathcal{J\&F} \end{math} on MeVIS. This demonstrates that the proposed SlowFast token design effectively unifies both tasks and enables coherent spatial–temporal understanding within a single framework.

\definecolor{lightgray}{HTML}{D9D9D9}
\begin{table*}[t]
\centering
\small
\setlength{\tabcolsep}{3.7pt} 
\renewcommand{\arraystretch}{1.1}
\begin{tabularx}{\linewidth}{l|cc|c|c|c|c|c|c|c|c >{\centering\arraybackslash}X}
\toprule
\multirow{2}{*}{\textbf{Dataset}} & \multicolumn{2}{c|}{\textbf{TVG}} & \textbf{VHD} & \textbf{YTVOS} & \textbf{ReVOS} & \textbf{VMME} & \textbf{MME} & \textbf{MMBench} & \textbf{SEED}  & \multicolumn{2}{c}{\textbf{LoomBench}} \\
\cmidrule(r){2-3} \cmidrule(r){4-4} \cmidrule(r){5-5} \cmidrule(r){6-6} \cmidrule(r){7-7} \cmidrule(r){8-8} \cmidrule(r){9-9} \cmidrule(r){10-10} \cmidrule(r){11-12} 
 & R1@0.5    &         mIoU             & mAP           & \begin{math} \mathcal{J\&F} \end{math}     & \begin{math} \mathcal{J\&F} \end{math}  & Acc      & P./R.     & Acc   &  Acc   & tIoU          & \begin{math} \mathcal{J\&F}_{bi\text{-}fore} \end{math}          \\
\midrule
Baseline & 66.2 & 56.5 & 25.8 & 70.0 & 60.0 & 50.7 & 492/115 & 79.0 & 73.9 & 28.1 & 34.6 \\
+VQA & 66.3 & 56.8 & 26.0 & 70.3 & 59.9 & 54.2 & 1684/623 & 80.9 & 74.7 & 29.8 & 36.9 \\
\rowcolor{lightgray}
+LoomData & \textbf{67.8} & \textbf{57.4} & \textbf{26.3} & \textbf{70.3} & \textbf{60.6} & \textbf{54.7} & \textbf{1699}/\textbf{628} & \textbf{81.1} & \textbf{75.0} & \textbf{34.8} & \textbf{41.9} \\
\bottomrule
\end{tabularx}
\vspace{-0.06in}
\caption{Ablation experiments on Training data.}
\label{tab:loomdata_results}
\vspace{-0.15in}
\end{table*}

\noindent \textbf{Effects of LoomData-8.7K.} Tab.~\ref{tab:loomdata_results} demonstrates the effectiveness of LoomData-8.7K in improving spatial–temporal understanding. We use VideoLoom trained on existing spatial and temporal datasets as the baseline. To eliminate the influence of additional VQA data, we include them only for a fair comparison. The results show that with LoomData-8.7K, our model achieves an improvement of +5.0 \begin{math} \mathcal{J\&F}_{bi\text{-}fore} \end{math} in joint spatial–temporal understanding, along with consistent gains across all benchmarks, including spatial, temporal, and general visual comprehension (VideoMME~\cite{fu2025video}, MME~\cite{fu2025mme}, MMBench~\cite{liu2024mmbench}, and SEED-Bench~\cite{li2024seed}). These results demonstrate that LoomData-8.7K could provide high-quality supervision for joint spatial–temporal understanding, with consistent spatial trajectories and temporal annotations.

\noindent \textbf{Effects of Base Models.} To evaluate the impact of different base models on spatial–temporal understanding, we conduct experiments using various MLLMs as backbones. As shown in Tab.~\ref{tab:backbone_results}, VideoLoom achieves higher performance with InternVL2.5-8B~\cite{chen2024expanding} compared to its smaller InternVL2.5-4B counterpart, indicating that 
\makebox[\linewidth][s]{larger language–vision models provide stronger multimodal representations for spatial–temporal reasoning.}
\vspace{-8pt}
\begin{table}[h!]  
    \begin{minipage}{0.45\columnwidth}
           When equipped with the more advanced InternVL3-8B~\cite{zhu2025internvl3}, further improvements are observed under comparable model capacities. These results demonstrate that VideoLoom continues to benefit from advancements in underlying MLLMs, showing strong scalability and the potential for even better spatial–temporal understanding as foundation models evolve.
    \end{minipage}
    \hfill  
    \begin{minipage}{0.53\columnwidth}
        \centering
        \small
        \setlength{\tabcolsep}{3.0pt} 
        \renewcommand{\arraystretch}{1.1}
        \begin{tabularx}{\linewidth}{l|c|c|c|c >{\centering\arraybackslash}X}
        \toprule
        \multirow{2}{*}{\textbf{Backbone}} & \textbf{TVG} & \textbf{VHD} & \textbf{ReVOS} & \multicolumn{2}{c}{\textbf{LoomBench}} \\
        \cmidrule(r){2-2} \cmidrule(r){3-3} \cmidrule(r){4-4} \cmidrule(r){5-6}       
                                 & mIoU          & mAP             & \begin{math} \mathcal{J\&F} \end{math}  & tIoU          & \begin{math} \mathcal{J\&F}_{bi\text{-}fore} \end{math}          \\
        \midrule
        InternVL2.5-4B~\cite{chen2024expanding} & 57.4 & 26.3 & 60.6 & 34.8 & 41.9 \\
        InternVL2.5-8B~\cite{chen2024expanding} & 56.4 & 27.1 & 62.0 & 40.2 & 47.2 \\
        InternVL3-8B~\cite{zhu2025internvl3} & 59.8 & 27.5 & 63.1 & 41.6 & 49.1 \\
        \bottomrule
        \end{tabularx}
        \vspace{-0.06in}
        \caption{Ablation experiments on Model size and type.}
        \label{tab:backbone_results}
    \end{minipage}
\vspace{-8pt}
\end{table}

\vspace{-0.1in}
\subsection{Visualizations}
\begin{figure*}[ht!]
\centering
\includegraphics[width=\textwidth]{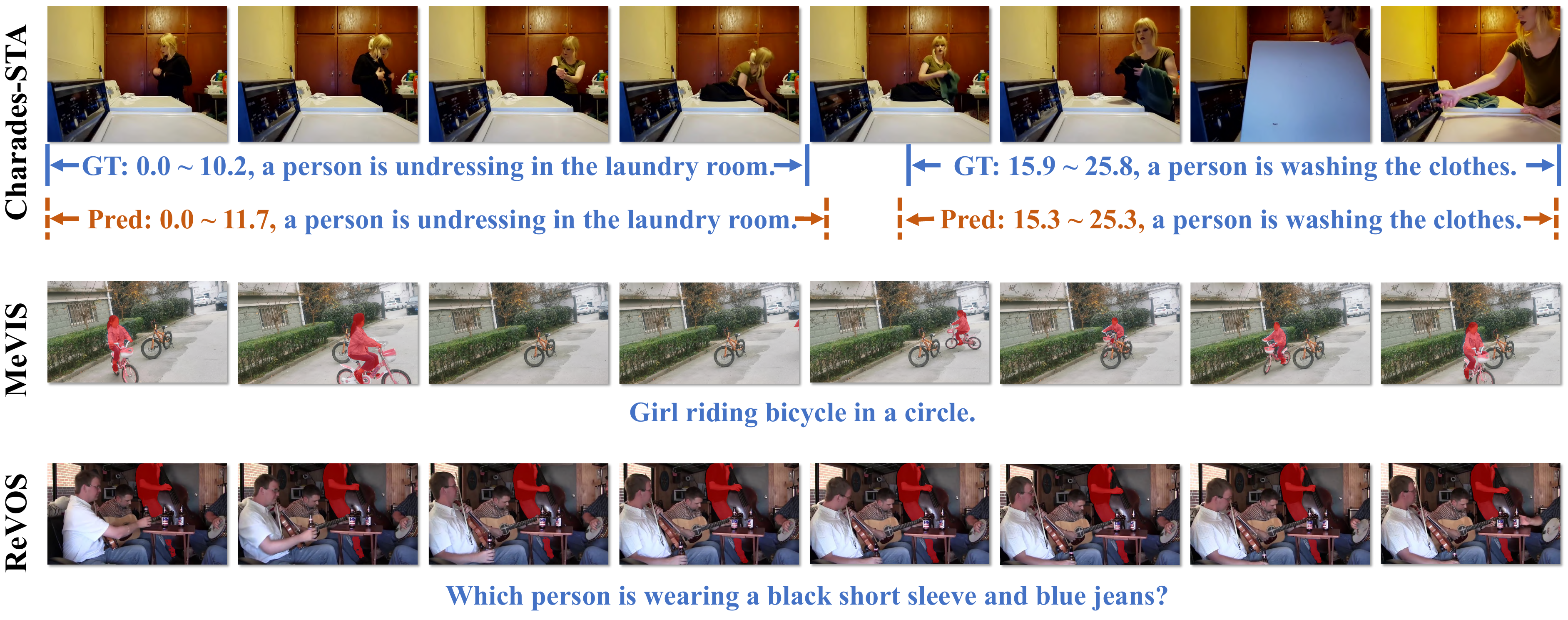}
\vspace{-0.25in}
\caption{Visualization of the predictions by VideoLoom on different spatial-temporal understanding tasks. From top to down, we show the visualization results of video temporal grounding on Charades-STA~\cite{gao2017tall}, referring VOS on MeVIS~\cite{ding2023mevis}, and reasoning VOS on ReVOS~\cite{yan2024visa}.}
\label{fig:visualization_public}
\vspace{-0.05in}
\end{figure*}

We present qualitative visualizations of VideoLoom across multiple spatial–temporal understanding datasets in Fig.~\ref{fig:visualization_public}. The first row illustrates that our model accurately localizes events along the temporal dimension, demonstrating its superior temporal modeling. The following two rows show its capability to perform object segmentation conditioned on diverse types of textual references (\eg, concise descriptions, reasoning-based queries). 

\begin{figure*}[h!]
\centering
\includegraphics[width=\textwidth]{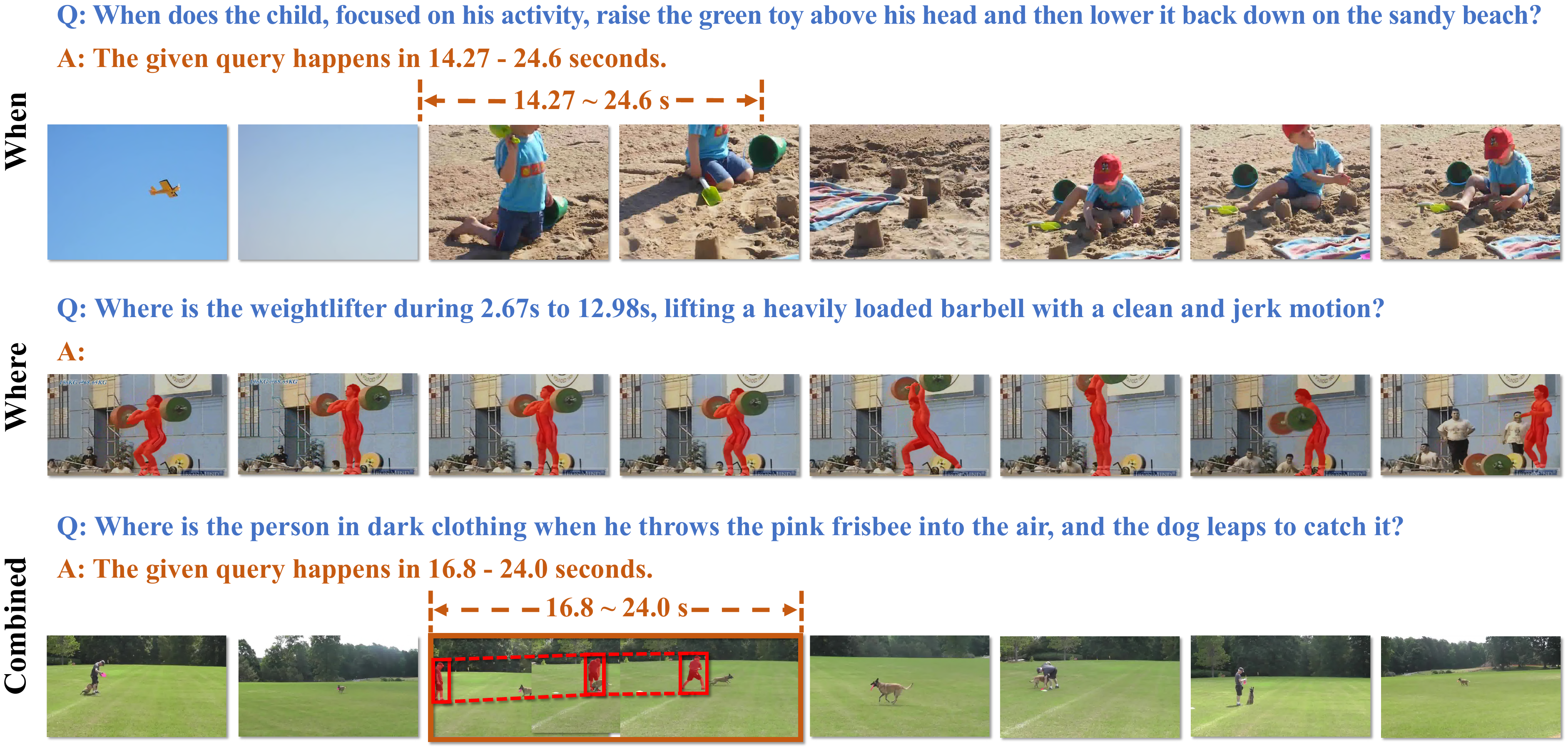}
\vspace{-0.25in}
\caption{Visualization of VideoLoom on LoomBench for \textit{When}, \textit{Where}, and \textit{Combined} questions.}
\label{fig:visualization_loombench}
\end{figure*}

Additionally, Fig.~\ref{fig:visualization_loombench} provides qualitative examples across the three question types from LoomBench, further illustrating the strong joint spatial–temporal understanding capability of VideoLoom. For instance, in the query ``Where is the person in dark clothing when he throws the pink frisbee into the air, and the dog leaps to catch it'', VideoLoom first localizes the relevant temporal segment corresponding to the throwing action and then accurately identifies the spatial region of the person within that interval. This example demonstrates its ability to reason across both time and space, linking dynamic actions to precise spatial localization within a unified framework.

\vspace{-0.05in}
\section{Conclusion}
\label{sec:conclusion}
\vspace{-0.02in}
This work presents the VideoLoom suite to advance joint spatial-temporal understanding. It comprises three key components: 1) LoomData-8.7k, a human-centric dataset that provides both timestamp-aligned action descriptions and fine-grained spatial masks. 2) VideoLoom, a unified Video LLM equipped with MLLM-SAM2 architecture to generate both temporal locations and spatial masks. and 3) LoomBench, a novel benchmark designed to evaluate Video LLMs across diverse question types, \textit{When}, \textit{Where}, and \textit{Combined}, for a comprehensive assessment of spatial-temporal understanding. Extensive experiments on a range of spatial and temporal benchmarks demonstrate that VideoLoom achieves strong performance and establishes new state-of-the-art results across multiple tasks. 

While already significantly reducing manual effort and enabling scalable annotation, the proposed annotation pipeline still involves multiple stages with interdependent components. In the future, we plan to further automate this process by integrating stronger multimodal foundation models and agents for both annotation generation and verification, aiming to further improve the efficiency and reliability.

\bibliographystyle{plainnat}
\bibliography{main}

\appendix
\clearpage
\section*{Appendix}

\section{Overview}
\vspace{-0.05in}

Our supplementary includes the following sections:
\begin{itemize}[itemindent=2em, leftmargin=0em]
\item \textbf{Sec.~\ref{section:model_details}: Model details.} Details for VideoLoom design, implementation and training data.
\item \textbf{Sec.~\ref{section:loomdata_details}: LoomData details.} Details for manual verification, and statistics for LoomData-8.7k.
\item \textbf{Sec.~\ref{section:experiment_results}: More experiment results.} Analysis on Bidirectional Foreground $\mathcal{J\&F}$, and additional performance evaluation.
\item \textbf{Sec.~\ref{section:more_visualization}: More visualization.} More visualization of our dataset and results.
\item \textbf{Sec.~\ref{section:prompt_design}: Prompt design.} Prompt for temporal action annotation and LoomBench construction.
\end{itemize}

\section{Model Details}
\label{section:model_details}

\vspace{-0.05in}
\subsection{More Details about VideoLoom}

\noindent\textbf{Interleaved Input.}
For temporal modeling, we interleave temporal information, \ie, unique frame IDs, with fast visual tokens. Specifically, we insert frame IDs, \eg, "This sampled frame id is 26", after the fast tokens of the corresponding frames, leading to an interleaved sequence. We then concatenate this token sequence with the slow tokens as input $I$ to the LLM:
\begin{equation}
I = [\mathrm{F}_1;\mathrm{ID}_1;...;\mathrm{F}_{N_{f}};\mathrm{ID}_{N_{f}};\mathrm{S_1};...;\mathrm{S}_{N_{s}}].
\end{equation}
where $\mathrm{ID}_j$, $\mathrm{F}_j$, $\mathrm{S}_k$ denote the ID text tokens, fast tokens, and slow tokens, while $N_{f}$ and $N_{s}$ for the count of frames with fast and slow tokens. 

By directly using numerical text of frame IDs to represent temporal positions, temporal understanding is transformed into language instruction QA, aligning with the general capabilities of MLLMs.

\noindent\textbf{[SEG] token.}
To generate masks for keyframes, SAM2~\cite{ravi2024sam} only needs to activate a visual encoder and a mask decoder. Given the keyframes sampled for slow tokens, we extract visual features $f_v$ using the visual encoder, which provides pixel-level details for trajectory prediction. The SAM2 mask decoder is connected to MLLM via a \texttt{[SEG]} token contained in the text output. Since MLLM performs fine-grained spatial-temporal modeling with SlowFast tokens, the \texttt{[SEG]} token captures rich target information under segmentation queries. The hidden states of the \texttt{[SEG]} token, denoted as $h_{seg}$, pass through an MLP projection layer to form a target embedding. This embedding serves as a novel visual prompt for SAM2, fed into the mask decoder with the visual features $f_v$ to generate masks $M_v$ for the keyframes:
\begin{equation}
M_v = \mathrm{SAM2}(f_v, \mathrm{MLP}(h_{\text{seg}})).
\end{equation}

\begin{table*}[t!]
\centering
\begin{minipage}{0.46\linewidth}
\centering
\small
\setlength{\tabcolsep}{8pt}
\renewcommand{\arraystretch}{1.3}
\begin{tabularx}{\linewidth}{l >{\centering\arraybackslash}X}
\toprule
\textbf{Hyperparameter} & \textbf{Value}        \\
\midrule
Epochs                  & 1            \\
Batch size              & 64           \\
Learning rate           & 4e-5         \\
Weight decay            & 0.05         \\
AdamW $\beta$           & (0.9, 0.999) \\
\midrule
Max sequence length for MLLM      & 8192        \\
Number of fast tokens per frame   & 16          \\
Number of slow tokens per frame   & 256         \\
Frame resolution for MLLM         & 448 $\times$ 448    \\
Frame resolution for SAM2         & 1024 $\times$ 1024  \\
\midrule
Number of frames for fast tokens  & $\leq$ 128  \\
Number of frames for slow tokens  & 5           \\
\bottomrule
\end{tabularx}
\vspace{-0.06in}
\caption{Hyperparameters for one-stage tuning.}
\label{tab:parameters}
\end{minipage}\hfill
\begin{minipage}{0.52\linewidth}
\centering
\small
\setlength{\tabcolsep}{6.5pt}
\renewcommand{\arraystretch}{1.1}
\captionsetup{width=\linewidth}
\begin{tabularx}{\linewidth}{l c >{\centering\arraybackslash}X}
\toprule
\textbf{Dataset} & \textbf{Item count} & \textbf{Repeats} \\
\midrule
LLaVA~\cite{liu2024improved}                       & 665K  & 1  \\
\midrule
RefCOCO~\cite{kazemzadeh2014referitgame}           & 17K   & 4  \\
RefCOCO+~\cite{kazemzadeh2014referitgame}          & 17K   & 4  \\
RefCOCOg~\cite{yu2016modeling}                     & 17K   & 4  \\
Grand-f~\cite{rasheed2024glamm} (Auto Annotated)   & 196K  & 1  \\
Grand-f~\cite{rasheed2024glamm} (Human Annotated)  & 1K    & 10 \\
\midrule
Charades-STA~\cite{gao2017tall}                    & 12.4K & 4  \\
YouCook2~\cite{zhou2018towards}                    & 1.2K  & 10 \\
QVHighlights~\cite{lei2021detecting}               & 6.9K  & 4  \\
\textbf{LoomData for VTG}                          & 8.7K  & 4  \\
\midrule
Ref-YTVOS~\cite{seo2020urvos}                      & 3.5K  & 12 \\
MeVIS~\cite{ding2023mevis}                         & 1.6K  & 12 \\
ReVOS~\cite{yan2024visa}                           & 1.7K  & 12 \\
\textbf{LoomData for refVOS}                       & 8.7K  & 4  \\
\bottomrule
\end{tabularx}
\vspace{-0.06in}
\caption{Training datasets, item counts, and repeat times.}
\label{tab:datasets}
\end{minipage}
\end{table*}

\subsection{Additional Implemental Details}

Tab.~\ref{tab:parameters} lists hyperparameters for one-stage tuning. Specifically, for the number of frames for fast tokens, we adopt different settings across datasets based on video duration, with a maximum of 128 frames. For Charades-STA~\cite{gao2017tall}, where videos typically last around 30 seconds, we sample 64 frames for fast tokens. For YouCook2~\cite{zhou2018towards}, where videos often exceed 2 minutes in length, we uniformly sample 128 frames. For QVHighlights~\cite{lei2021detecting}, annotated in 2-second intervals, we sample frames at 2 FPS, typically yielding around 75 frames. For spatial datasets~\cite{seo2020urvos,ding2023mevis,yan2024visa}, which provide annotated frame sequences, we uniformly sample up to 64 frames.

\vspace{-0.05in}
\subsection{Training Data}

We present all the datasets for training and report their item counts and repeat times in Tab.~\ref{tab:datasets}. Finally, VideoLoom is jointly trained for 1,315K iterations and achieves advanced performance on all these tasks.

\section{LoomData Details}
\label{section:loomdata_details}

\subsection{Details about Manual Verification}

\begin{figure*}[ht!]
\centering
\includegraphics[width=\textwidth]{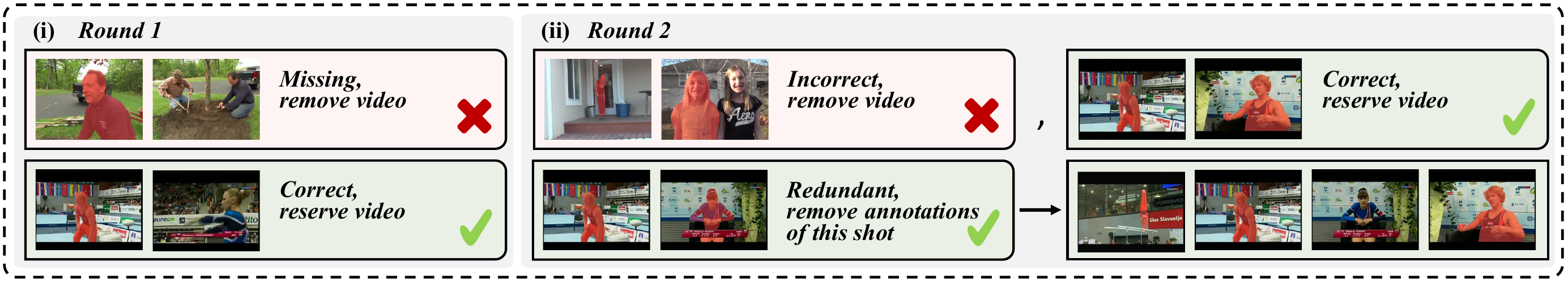}
\vspace{-0.2in}
\caption{Examples of manual verification. (i) In the first round, we filter out videos with missing annotations. (ii) In the second round, we filter out videos with incorrect annotations and remove redundant annotations from the shots of the retained videos.}
\label{fig:manual}
\vspace{-0.15in}
\end{figure*}

Here we introduce the simple manual verification process in the pipeline, explaining how to implement filtering and correction of complete tracklets after spatial mask annotation. This approach involves two rounds of simple judgments to minimize manual involvement:

In the first round, we primarily focus on filtering out videos with missing annotations, as completing the missing tracklets requires extensive manual annotation, which is not scalable. Specifically, we display the annotation on the middle frame of the longest shot (\ie, the key frame where we initially identify the main character) as a reference, and then display the middle frames of the shots without tracklets in turn. We then manually determine whether there is an unlabeled main character in these frames, discarding the video if one is found, as shown in Fig.~\ref{fig:manual}. (i).

The second round of verification focuses on the shots with tracklets, where we filter out videos with incorrect annotations and remove redundant annotations from the retained shots. For a shot with tracklets, we define incorrect annotations as the presence of the main character but the mask labeled to other objects, and redundant annotations as the absence of the main character but the mask labeled to other objects. We discard entire videos containing incorrectly labeled shots and remove annotations from redundantly labeled shots to make a simple revision of the video, as shown in Fig.~\ref{fig:manual}. (ii). Specifically, we continue to display the annotation of the middle frame of the longest shot for reference purposes and display the middle frames of the shots with tracklets in turn. We then manually determine whether the annotations on these frames are incorrect, redundant, or correct to carry out the corresponding operations.

\subsection{Statistics for LoomData-8.7k}
\label{section:statistic_loomdata}

\begin{figure*}[h!]
\centering
\includegraphics[width=\textwidth]{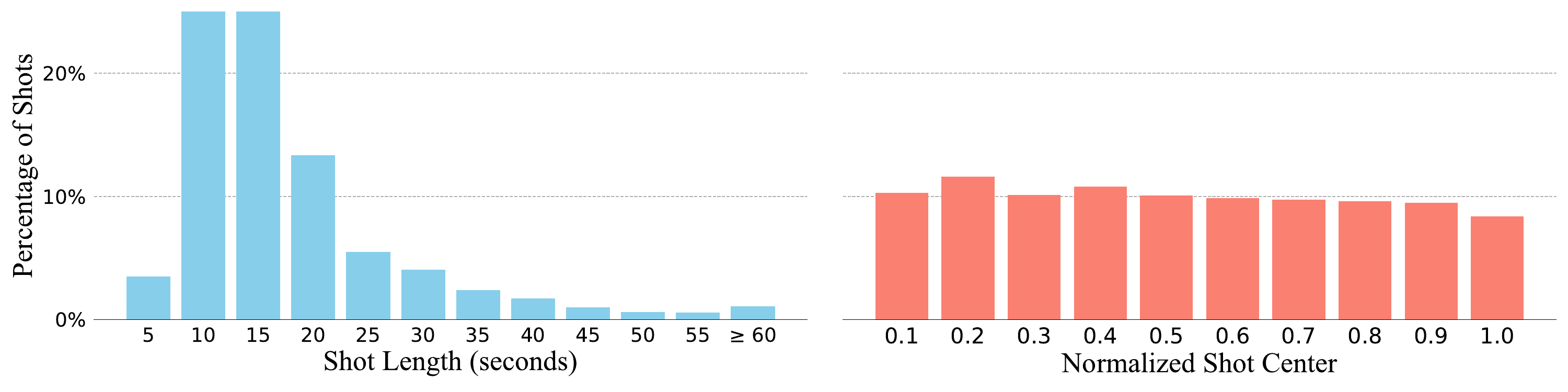}
\vspace{-0.25in}
\caption{Distribution of shot lengths (left) and normalized (by video duration) center timestamps (right). The shots vary widely in length, and they distribute almost evenly along the videos.}
\label{fig:loomdata}
\end{figure*}

Tab.~\ref{tab:datasets_comparison} compares our constructed dataset, LoomData-8.7k, with existing spatial-temporal datasets. For the first time, LoomData achieves joint annotation of temporal timestamps and spatial masks on nearly 2-minute videos. LoomData enables fine-grained temporal partition, with each video containing an average of 6.0 segments with tracklets, comparable to current spatial-temporal datasets. Compared to temporal datasets, which only roughly label overlapping temporal locations, LoomData achieves a complete temporal partition of the videos while providing a more detailed description. Compared to spatial datasets, LoomData achieves mask-level annotation while ensuring instance consistency across the entire video. Fig.~\ref{fig:loomdata} shows the distribution of shot lengths and normalized shot center timestamps (by video duration). LoomData contains shots of widely varying lengths. Over 50\% of shots are concentrated in the range from 5 to 15 seconds, while only a few exceed 30 seconds. These shots are almost evenly distributed across the videos, suggesting that LoomData suffers less from temporal bias.

\begin{table*}[!h]
\centering
\small
\setlength{\tabcolsep}{3.5pt} 
\renewcommand{\arraystretch}{1.1}
\begin{tabularx}{\linewidth}{l|>{\centering\arraybackslash}X >{\centering\arraybackslash}X >{\centering\arraybackslash}X cccc}
\toprule
\textbf{Dataset}           & \textbf{\#Videos} & \begin{tabular}[c]{@{}c@{}}\textbf{Avg} \\ \textbf{\#Segments}\end{tabular} & \begin{tabular}[c]{@{}c@{}}\textbf{Avg} \\ \textbf{\#Tracklets}\end{tabular} & \begin{tabular}[c]{@{}c@{}}\textbf{Avg Len (sec)} \\ \textbf{Segment/Video}\end{tabular} & \textbf{Temporal Ann.} & \textbf{Box Ann.} & \textbf{Mask Ann.} \\
\midrule
Charades-STA~\cite{gao2017tall}      & 5,338    & 6.8            & -               & 8.1/30.6                    & \ding{51}             &          &           \\
ANet Captions~\cite{krishna2017dense} & 10,024   & 3.7            & -               & 36.2/117.6                  & \ding{51}             &          &           \\
RefYTVOS~\cite{seo2020urvos}          & 3,471    & -              & 1.9             & -                           &               &          & \ding{51}         \\
MeVIS~\cite{ding2023mevis}             & 1,662    & -              & 4.3             & -                           &               &          & \ding{51}         \\
VidSTG~\cite{zhang2020does}            & 5,563    & 6.5            & 5.0             & 9.7/28.0                    & \ding{51}             & \ding{51}        &           \\
\midrule
LoomData          & 1,456    & 6.0            & 6.0             & 15.0/102.2                  & \ding{51}             &          & \ding{51}         \\
\bottomrule
\end{tabularx}
\vspace{-0.06in}
\caption{Comparison with existing spatial-temporal datasets.}
\label{tab:datasets_comparison}
\vspace{-0.1in}
\end{table*}

\section{More Experiment Results}
\label{section:experiment_results}

\subsection{Analysis on Bidirectional Foreground J\&F}
\label{section:bf_jf}

We propose a new evaluation metric, Bidirectional Foreground \begin{math} \mathcal{J\&F} \end{math}, for assessing the joint spatial-temporal understanding of \textit{Combined} questions on LoomBench. In this section, we first demonstrate the necessity with experimental results under varying queried segment lengths. We then present the specific values of each component of this metric to provide an in-depth assessment.

\begin{table}[h!]  
    \begin{minipage}{0.45\columnwidth}
           \noindent\textbf{The Necessity of \begin{math} \mathcal{J\&F}_{bi\text{-}fore} \end{math}.} The \textit{Combined} questions are divided by the percentage of length of the queried segment over the entire video, into three categories: 0-20\%, 20-60\%, and 60-100\%. We then provide a comparative analysis of the standard \begin{math} \mathcal{J\&F} \end{math} and our proposed \begin{math} \mathcal{J\&F}_{bi\text{-}fore} \end{math} in Tab.~\ref{tab:combined_results_under_lengths}.
    \end{minipage}
    \hfill  
    \begin{minipage}{0.53\columnwidth}
        \centering
        \small
        \setlength{\tabcolsep}{8.5pt} 
        \renewcommand{\arraystretch}{1.1}
        \begin{tabularx}{\columnwidth}{l|c c >{\centering\arraybackslash}X |c}
        \toprule
        \multirow{1}{*}{\textbf{Metric}} & \textbf{0-20\%} & \textbf{20-60\%} & \textbf{60-100\%} & \textbf{All} \\
        \midrule
        Standard \begin{math} \mathcal{J\&F} \end{math}    & 88.9 & 77.7 & 41.0   & 83.3          \\
        \begin{math} \mathcal{J\&F}_{bi\text{-}fore} \end{math}            & 47.6 & 50.8 & 37.1   & 49.1 \\
        \bottomrule
        \end{tabularx}
        \vspace{-0.06in}
        \caption{Comparison between standard \begin{math} \mathcal{J\&F} \end{math} and \begin{math} \mathcal{J\&F}_{bi\text{-}fore} \end{math} on \textit{Combined} questions of LoomBench, under varying queried segment lengths.}
        \label{tab:combined_results_under_lengths}
    \end{minipage}
\vspace{-8pt}
\end{table}

We can see that \begin{math} \mathcal{J\&F}_{bi\text{-}fore} \end{math} performs stably under variable-length queried segments, while the standard \begin{math} \mathcal{J\&F} \end{math} increases significantly with shorter lengths, resulting in a substantial gap between 0-20\% and 60-100\%. However, VideoLoom does not demonstrate superiority in short segments. On the contrary, it is significantly more challenging to perform spatial-temporal localization for short segments over the whole video. This is due to the calculation of \begin{math} \mathcal{J\&F} \end{math}. For segments without masklets, \ie, background segments, when the predicted mask is None, the value of \begin{math} \mathcal{J\&F} \end{math} reaches 1 (100\%). When computed over the entire video, \begin{math} \mathcal{J\&F} \end{math} is significantly influenced by the easily predicted background segments, leading to inflated values and excessive sensitivity to the proportion of foreground queries, which prevents a correct assessment of spatial-temporal capabilities.

Referring VOS~\cite{seo2020urvos,ding2023mevis,yan2024visa} adopts standard \begin{math} \mathcal{J\&F} \end{math} as evaluation metrics because videos in existing datasets are often foreground throughout (up to 60\% or more, as indicated by Tab.~\ref{tab:combined_results_under_lengths} showing close values for the two metrics on segments with length of 60-100\%), which is notably different from LoomBench. To effectively evaluate performance on LoomBench, we utilize the Bidirectional Foreground \begin{math} \mathcal{J\&F} \end{math} metric, thereby avoiding extensive computation on background segments and ensuring accurate assessment for spatial-temporal comprehension. 

\noindent\textbf{In-depth Comparison of Components.} We present the specific values of each component of \begin{math} \mathcal{J\&F}_{bi\text{-}fore} \end{math} in Tab.~\ref{tab:combined_details}, including \begin{math} \mathcal{J}_p \end{math}, \begin{math} \mathcal{F}_p \end{math}, \begin{math} \mathcal{J\&F}_p \end{math} computed over the predicted masklet, and \begin{math} \mathcal{J}_g \end{math}, \begin{math} \mathcal{F}_g \end{math}, \begin{math} \mathcal{J\&F}_g \end{math} computed over the groundtruth. The experimental results demonstrate that VideoLoom outperforms the baseline, which consists of TimeSuite~\cite{zeng2024timesuite} and Sa2VA~\cite{yuan2025sa2va}, across all metrics. Additionally, it is evident that the metric scores computed over the predicted masklet are higher than those computed over the groundtruth, highlighting the superior precision of the model predictions, though a notable gap remains in recall.

\begin{table*}[!ht]
\centering
\small
\setlength{\tabcolsep}{10pt} 
\renewcommand{\arraystretch}{1.1}
\begin{tabularx}{\linewidth}{l|>{\centering\arraybackslash}X >{\centering\arraybackslash}X >{\centering\arraybackslash}X| >{\centering\arraybackslash}X >{\centering\arraybackslash}X >{\centering\arraybackslash}X| >{\centering\arraybackslash}X}
\toprule
\multirow{1}{*}{\textbf{Method}}
                        & \textbf{\begin{math} \mathcal{J}_p \end{math}} & \textbf{\begin{math} \mathcal{F}_p \end{math}} & \textbf{\begin{math} \mathcal{J\&F}_p \end{math}} & \textbf{\begin{math} \mathcal{J}_g \end{math}} & \textbf{\begin{math} \mathcal{F}_g \end{math}} & \textbf{\begin{math} \mathcal{J\&F}_g \end{math}} & \textbf{\begin{math} \mathcal{J\&F}_{bi\text{-}fore} \end{math}} \\
\midrule
TimeSuite+Sa2VA    & 47.0 & 48.9 & 48.0    & 25.4 & 26.6 & 26.0    & 33.7          \\
VideoLoom-8B            & 58.1 & 60.5 & 59.3    & 41.1 & 42.8 & 41.9    & 49.1         \\
\bottomrule
\end{tabularx}
\vspace{-0.06in}
\caption{Detailed results of VideoLoom on \textit{Combined} questions of LoomBench.}
\label{tab:combined_details}
\end{table*}

\subsection{Ablation on Non-Human Categories}

To demonstrate the generalizability of VideoLoom on human and non-human categories, we conduct ablation experiments on RefDavis17~\cite{khoreva2019video}, a benchmark for referring VOS, in a zero-shot setting. We divide the classes of objects and report the results separately in Tab.~\ref{tab:non_human_results}.

With or without LoomData, the performance of segmentation on human class surpasses that of the non-human classes. Moreover, incorporating our constructed LoomData leads to a notable enhancement in the segmentation of the human class (+2.3 \begin{math} \mathcal{J\&F} \end{math}), while also benefiting the segmentation of non-human classes (+2.3 \begin{math} \mathcal{J\&F} \end{math}). This suggests that, although our data primarily targets the human class, detailed textual descriptions contribute to the comprehension of semantics across various categories. Consequently, VideoLoom demonstrates the ability to generalize to any category and greatly benefits from the human class annotations provided by LoomData.

\definecolor{lightgray}{HTML}{D9D9D9}
\begin{table*}[t]
\centering
\small
\setlength{\tabcolsep}{5.7pt} 
\renewcommand{\arraystretch}{1.1}
\begin{tabularx}{\linewidth}{l|>{\centering\arraybackslash}X>{\centering\arraybackslash}X>{\centering\arraybackslash}X|>{\centering\arraybackslash}X>{\centering\arraybackslash}X>{\centering\arraybackslash}X|>{\centering\arraybackslash}X>{\centering\arraybackslash}X>{\centering\arraybackslash}X}
\toprule
\multirow{2}{*}{\textbf{Method}} & \multicolumn{3}{c|}{\textbf{Human}} & \multicolumn{3}{c|}{\textbf{Non-Human}} & \multicolumn{3}{c}{\textbf{All}} \\
& \begin{math} \mathcal{J} \end{math}    &         \begin{math} \mathcal{F} \end{math}             & \begin{math} \mathcal{J\&F} \end{math}           & \begin{math} \mathcal{J} \end{math}     & \begin{math} \mathcal{F} \end{math}  & \begin{math} \mathcal{J\&F} \end{math}      & \begin{math} \mathcal{J} \end{math}     & \begin{math} \mathcal{F} \end{math}   &  \begin{math} \mathcal{J\&F} \end{math} \\
\midrule
w/o LoomData & 73.0 & 82.0 & 77.5 & 63.1 & 72.3 & 67.7 & 67.5 & 76.6 & 72.1 \\
Ours & 75.4 & 84.3 & 79.8 & 65.7 & 74.3 & 70.0 & 70.0 & 78.7 & 74.3 \\
\bottomrule
\end{tabularx}
\vspace{-0.06in}
\caption{Ablation experiments on Human and Non-human categories of RefDavis17~\cite{khoreva2019video}.}
\label{tab:non_human_results}
\vspace{-0.1in}
\end{table*}

\subsection{Detailed Comparison with Sa2VA}

\makebox[\linewidth][s]{We conduct a fair comparison with Sa2VA~\cite{yuan2025sa2va}, the model most closely aligned with our approach. Following}
\vspace{-8pt}
\begin{table}[h!] 
    \begin{minipage}{0.4\columnwidth}
            Sa2VA, we employ InternVL2.5-4B~\cite{chen2024expanding} as the MLLM backbone to report our results in Table~\ref{tab:sa2va_results}. We can see that VideoLoom significantly surpasses Sa2VA on MeVIS~\cite{ding2023mevis} and also achieves competitive performance on RefYTVOS~\cite{seo2020urvos}, highlighting its superior motion capture and reasoning capabilities.
    \end{minipage}
    \hfill  
    \begin{minipage}{0.58\columnwidth}
        \centering
        \small
        \setlength{\tabcolsep}{3.0pt} 
        \renewcommand{\arraystretch}{1.1}
        \setlength{\tabcolsep}{3.5pt}
        \begin{tabularx}{\columnwidth}{l| l| >{\centering\arraybackslash}X >{\centering\arraybackslash}X >{\centering\arraybackslash}X}
        \toprule
        \multirow{2}{*}{\textbf{Method}} & \multirow{2}{*}{\textbf{Backbone}} & \textbf{MeVIS\_u} & \textbf{MeVIS} & \textbf{YTVOS} \\ 
        \cmidrule(r){3-3} \cmidrule(r){4-4} \cmidrule(r){5-5}
        & & \begin{math} \mathcal{J\&F} \end{math} & \begin{math} \mathcal{J\&F} \end{math} & \begin{math} \mathcal{J\&F} \end{math} \\
        \midrule
        Sa2VA~\cite{yuan2025sa2va} & InternVL2.5-4B & 55.9 & 46.4 & 71.3 \\
        VideoLoom & InternVL2.5-4B & 60.9 & 50.6 & 70.3 \\
        \bottomrule
        \end{tabularx}
        \vspace{-0.06in}
        \caption{Comparison with Sa2VA~\cite{yuan2025sa2va} using the same backbone.}
        \label{tab:sa2va_results}
    \end{minipage}
\vspace{-8pt}
\end{table}

\section{More Visualization}
\label{section:more_visualization}

\subsection{Visualization of Full Annotation}

\begin{figure*}[ht!]
\centering
\includegraphics[width=\textwidth]{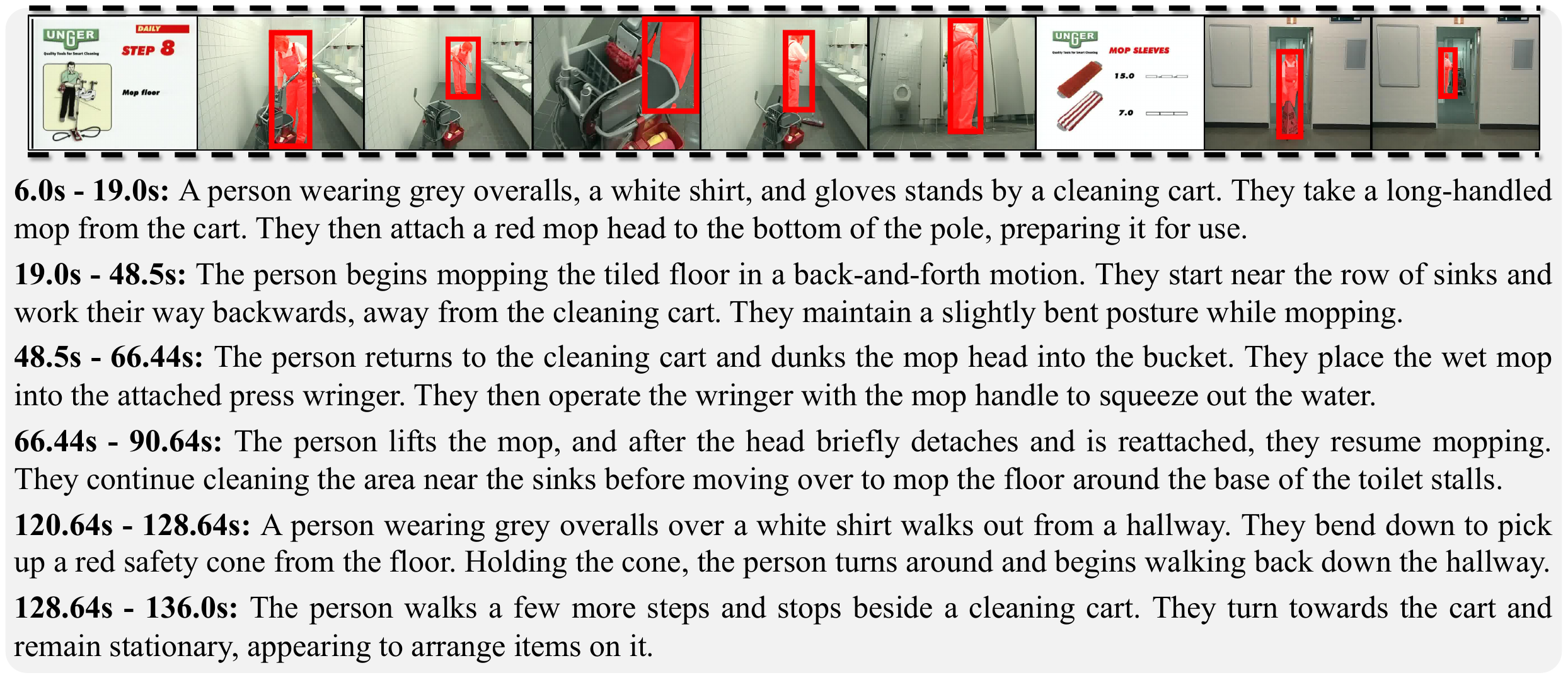}
\vspace{-0.2in}
\caption{An example of the complete spatial-temporal annotation of a video.}
\label{fig:loomdata_case}
\end{figure*}

To visualize the annotation results of our pipeline, we present an example of the complete spatial-temporal annotation for a randomly selected video in Fig.~\ref{fig:loomdata_case}. This annotation fully captures the timestamp-aligned actions and mask-level locations of the main characters.

\subsection{Qualitative Results and Failure Cases}
\begin{figure*}[ht!]
\centering
\includegraphics[width=0.92\textwidth]{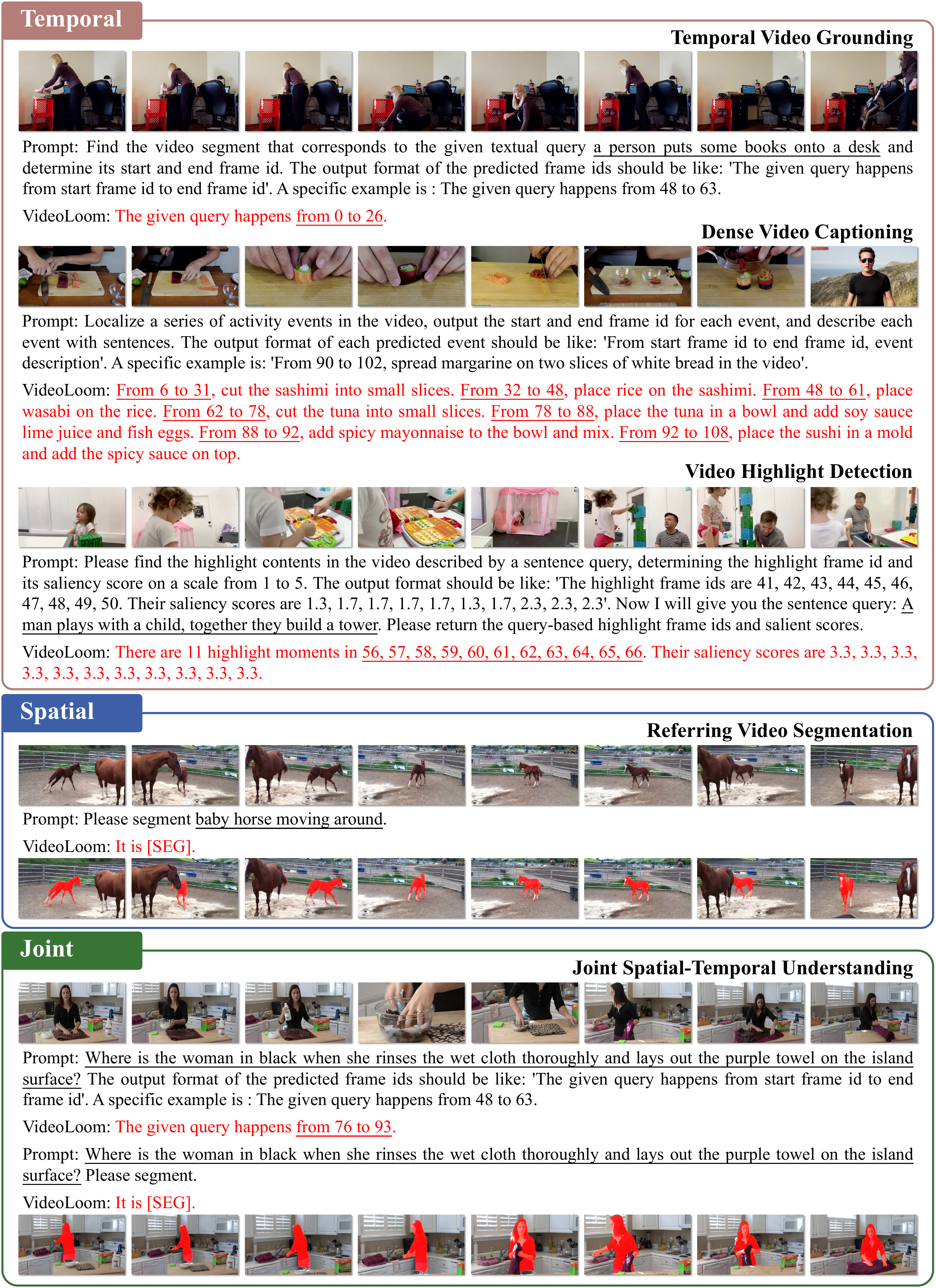}
\vspace{-0.1in}
\caption{Additional qualitative results of VideoLoom on diverse spatial-temporal tasks.}
\label{fig:additional_qualitative_results}
\vspace{-0.2in}
\end{figure*}

\begin{figure*}[ht!]
\centering
\includegraphics[width=\textwidth]{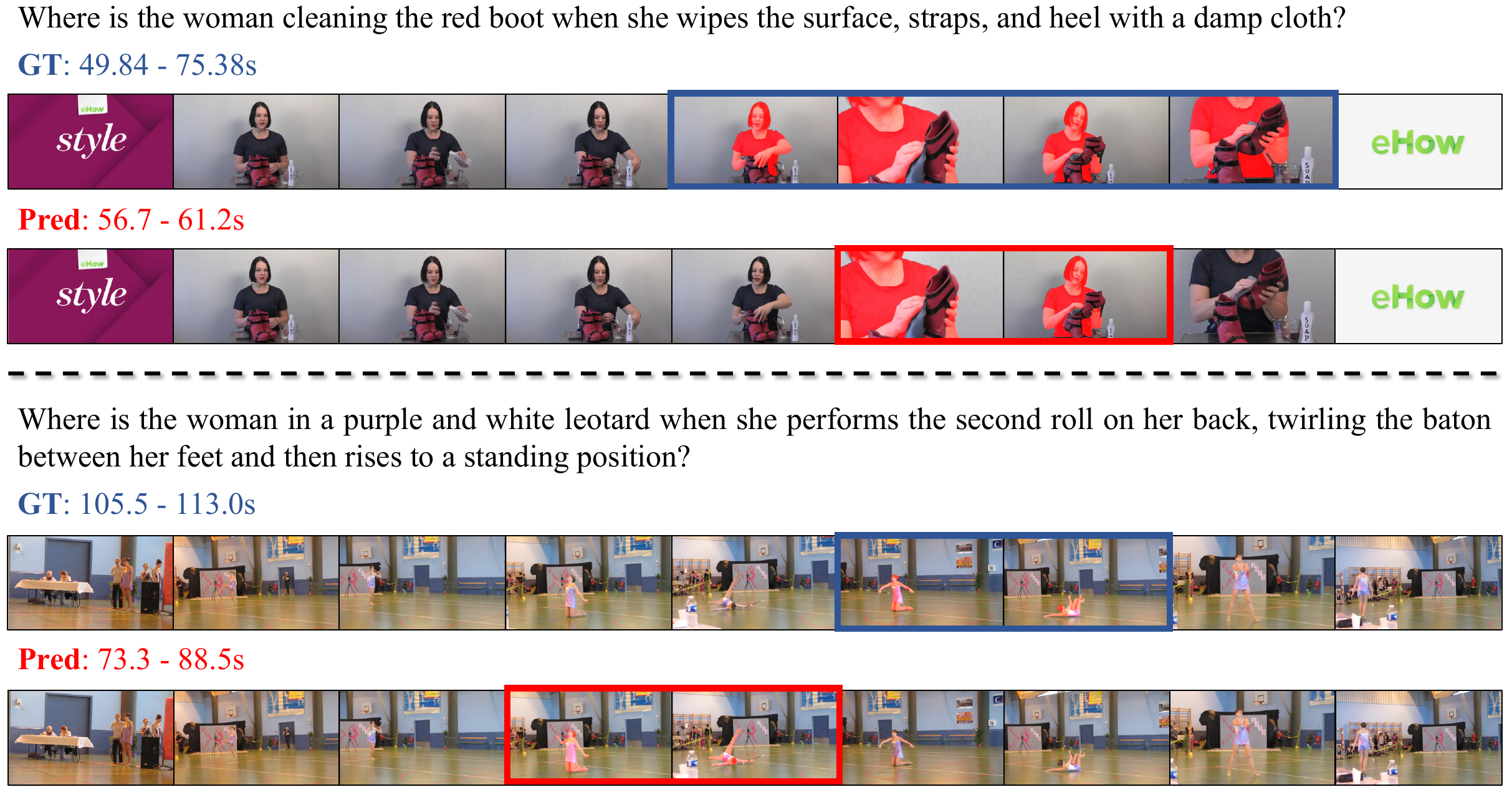}
\vspace{-0.1in}
\caption{Failure cases of VideoLoom on LoomBench, \eg, when querying sub-actions or the \textit{n}-th occurrence.}
\label{fig:failure_cases}
\end{figure*}

We present additional qualitative results of VideoLoom across multiple spatial-temporal tasks. As illustrated in Fig.~\ref{fig:additional_qualitative_results}, VideoLoom can follow diverse spatial-temporal instructions and establish a solid baseline across different tasks. However, in complex joint understanding scenarios (\eg, when querying sub-actions or the \textit{n}-th occurrence), it occasionally generates inaccurate spatial-temporal locations, as shown in Fig.~\ref{fig:failure_cases}. This issue likely arises from limitations in temporal action grounding. When confronted with lengthy queries, the model struggles to identify complete temporal intervals spanning the entire motion sequence, which may lead to misaligned spatial-temporal localization. We plan to explore this issue further in future work.

\section{Prompt Design}
\label{section:prompt_design}

\subsection{Prompt for Temporal Action Annotation}
\begin{figure*}[ht!]
\centering
\includegraphics[width=\textwidth]{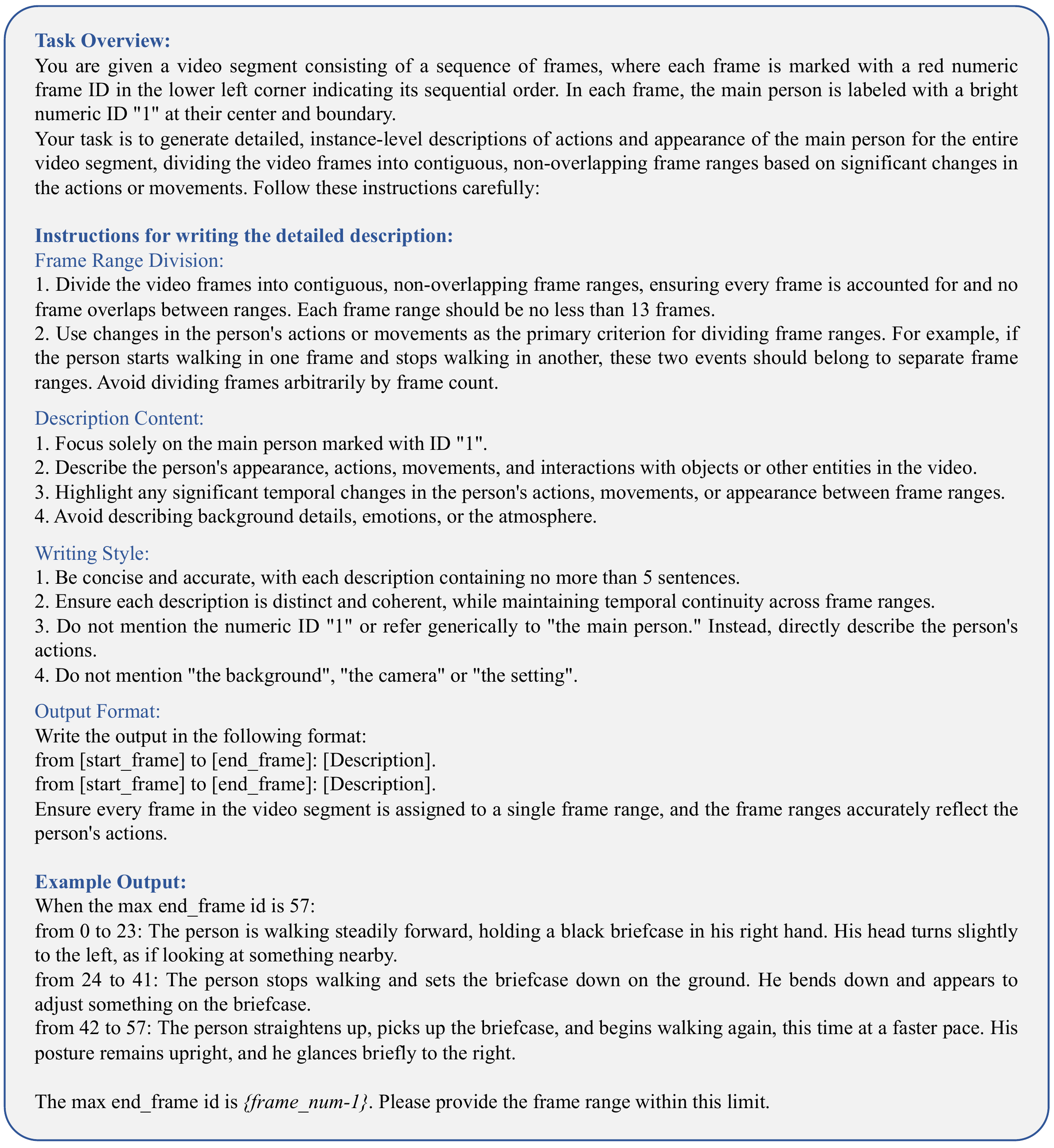}
\vspace{-0.2in}
\caption{Instruction format for guiding Gemini2.5pro~\cite{comanici2025gemini} to generate detailed and distinct action descriptions, the \textit{italicized} part are placeholders for the text inputs.}
\label{fig:prompt}
\end{figure*}

Prompt engineering plays a vital role in guiding Gemini2.5pro~\cite{comanici2025gemini} to generate detailed and specific action descriptions aligned with frame IDs for video shots. The prompt utilized is illustrated in Fig.~\ref{fig:prompt}. To ensure clarity and precision, we first outline the task of generating instance-level descriptions of actions and appearances using visual prompts from SoM~\cite{yang2023set} and NumPro~\cite{wu2025number}. The former annotates instance-level IDs on the main character, while the latter sequentially labels unique frame IDs on each frame. Next, we provide a series of instructions, including Frame Range Division, Description Content, Writing Style, and Output Format. These guidelines ensure concise, distinct, and formatted output with complete temporal coverage, avoiding irrelevant descriptions. Finally, we provide an example output and specify the number of sampled frames for the shot, ensuring alignment between the descriptions and the frame IDs. As a result, Gemini2.5pro generates clear and accurate instance-level descriptions of the main character based on the visual content of the current shot, under the guidance of our carefully designed prompt.

\subsection{Prompt for LoomBench Construction}
\begin{figure*}[ht!]
\centering
\includegraphics[width=\textwidth]{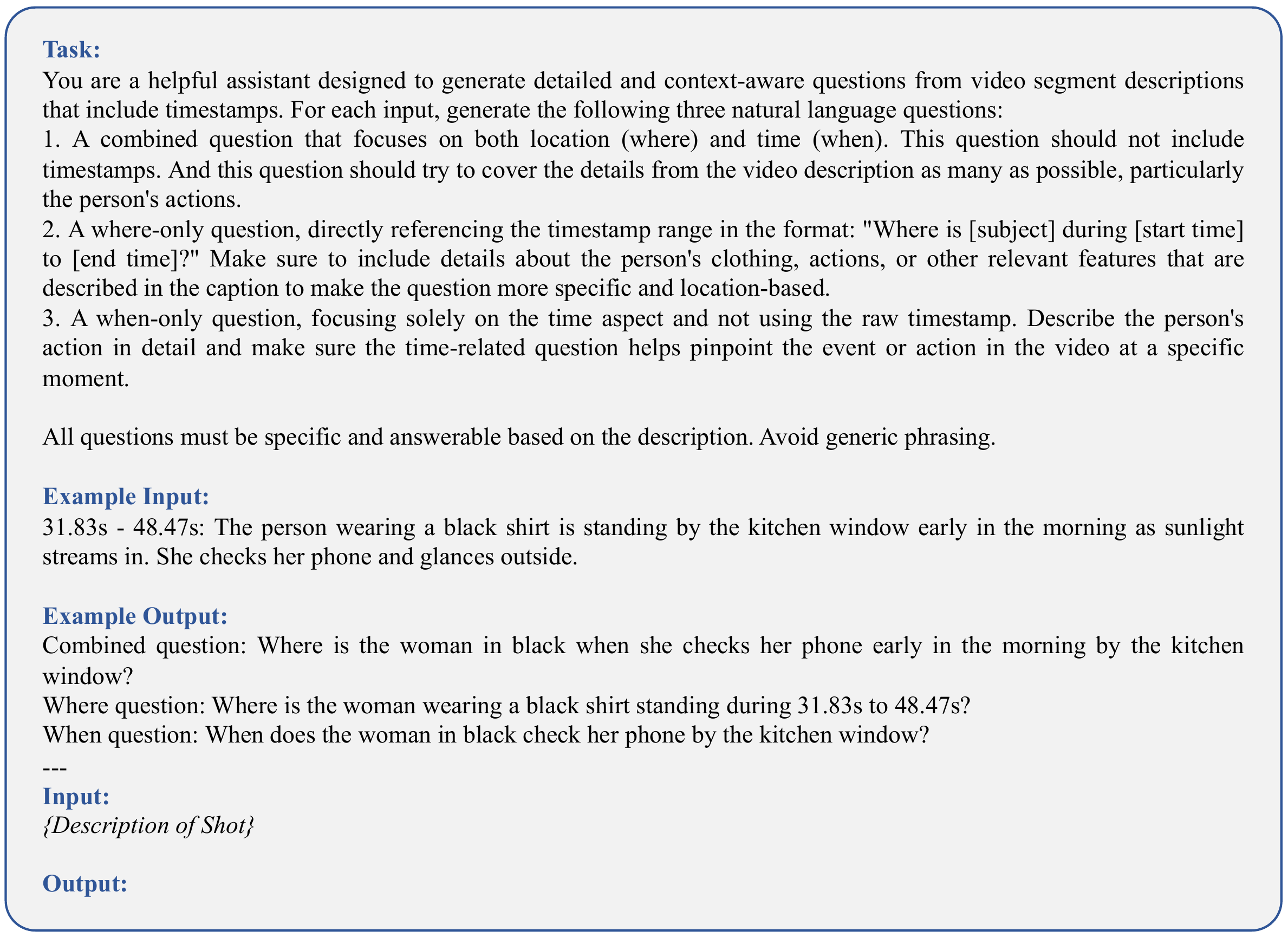}
\vspace{-0.2in}
\caption{Instruction format for guiding LLaMA3.1~\cite{grattafiori2024llama} to generate three types of questions to construct LoomBench, the \textit{italicized} part are placeholders for the text inputs.}
\label{fig:prompt_bench}
\end{figure*}

We prompt LLaMA3.1~\cite{grattafiori2024llama} to generate \textit{When}, \textit{Where}, and \textit{Combined} questions based on annotations produced by our pipeline, and we show the prompt in Fig.~\ref{fig:prompt_bench}. We first define the task to create detailed and context-aware questions from video shot descriptions explicitly. Next, we specify the requirements for each of the three question types, emphasizing that timestamps should not appear in \textit{Combined} or \textit{When} questions. Finally, we present a concrete example to clarify the form of the questions further. Based on each shot description, LLaMA3.1 subsequently generates three categories of questions, both detailed and context-aware, to construct the LoomBench.

\end{document}